\pdfoutput=1   %
\documentclass{article}
\usepackage{iclr2027_conference,times}
\usepackage[utf8]{inputenc}
\usepackage{textcomp}

\usepackage{amsmath,amsfonts,bm}

\def\eqref#1{equation~\ref{#1}}

\def\1{\bm{1}}

\DeclareMathAlphabet{\mathsfit}{\encodingdefault}{\sfdefault}{m}{sl}
\SetMathAlphabet{\mathsfit}{bold}{\encodingdefault}{\sfdefault}{bx}{n}

\usepackage{amsmath,amssymb,mathtools}
\usepackage{booktabs}
\usepackage{float}
\usepackage{graphicx}
\usepackage{wrapfig}
\makeatletter\providecommand\captionof[1]{\def\@captype{#1}\caption}\makeatother %
\IfFileExists{needspace.sty}{\usepackage{needspace}}{\providecommand{\Needspace}[1]{}}
\usepackage{microtype}
\usepackage{xspace}
\usepackage{multirow}
\usepackage{enumitem}
\usepackage{tabularx}
\usepackage{xcolor}
\usepackage{tikz}
\usetikzlibrary{arrows.meta,positioning,fit,calc,backgrounds,shapes.geometric}
\usepackage{url}
\usepackage{hyperref}
\hypersetup{hidelinks}

\title{Coarse Semantic Blocks for \\ Efficient Training-Free Sparse Attention}

\author{Sosuke Kobayashi, Shoichiro Yamaguchi, Seiya Tokui \\
Preferred Networks \\
\texttt{\{sosk,guguchi,tokui\}@preferred.jp}}
\iclrfinalcopy
\newcommand{\codeurl}{\url{https://github.com/pfnet-research/semantic-block-sparse-attention/}}

\begin{document}
\maketitle
\lhead{Preprint}   %
\suppressfloats[t] %

\begingroup
\makeatletter
\let\@footnotetext\H@@footnotetext
\makeatother
\renewcommand{\thefootnote}{}
\footnotetext{Code: \codeurl}
\endgroup

\begin{abstract}
We introduce a training-free sparse attention method that builds a prompt-specific key index to accelerate both prefill and decoding.
The method clusters the keys into large semantic blocks that group tokens from noncontiguous positions.
Each query position selects blocks by scoring their centroids and attends to the original keys and values within them.
Queries that select the same block are processed together, so prefill stays fast even though each query chooses its own blocks.
Large blocks make centroid scoring and this grouped processing cheap, so each query can attend to more tokens in the same time.
Larger blocks with a larger token budget can thus be both faster and more accurate on long-context tasks than smaller blocks with a smaller budget.
We evaluate two hybrid LLMs, Qwen3.5-4B and 35B-A3B, on one NVIDIA H200.
Including the cost of building and using the index, whole-model wall-clock speedups over dense attention are 2.3–2.4× for prefill and 1.7–2.0× for decoding at 512K tokens, and 2.9–3.0× for prefill at 1M.
The 35B-A3B model stays close to dense attention on five RULER tasks through 1M, on LongBench-v2 overall, and on MRCR through 256K.
The 4B model loses accuracy mainly on multi-key retrieval, yet outperforms dense Qwen3.5-2B, whose latency is similar at 512K, on all three benchmarks.

\end{abstract}

\section{Introduction}
\label{sec:introduction}
Long contexts make attention expensive during both prefill and decoding.
Sparse attention restricts each query to a subset of tokens, but it is faster only if choosing the tokens and preparing their keys and values costs less than the attention computation it saves.
Methods with learned indexers, such as DSA~\citep{deepseek2025v32}, MSA~\citep{lai2026msa}, and QSA~\citep{qwen2026qsa}, train the indexer and the model so that a small token budget suffices.
For training-free methods, estimating token importance separately for each query during prefill is costly, and irregular selections are difficult to execute efficiently on GPUs~\citep{nawrot2026sparsefrontier}.
Existing methods therefore select contiguous blocks~\citep{tang2024quest}, whose tokens are often unrelated to one another, or share one selection across neighboring queries~\citep{jiang2024minference,lai2025flexprefill,jones2026quoka,fan2026flashprefillv2}.
Other methods cluster similar keys during dense prefill and use the resulting index only for decoding~\citep{liu2025clusterkv,zhu2026tactic}.

\begin{figure}[t]
\centering
\includegraphics[width=\linewidth]{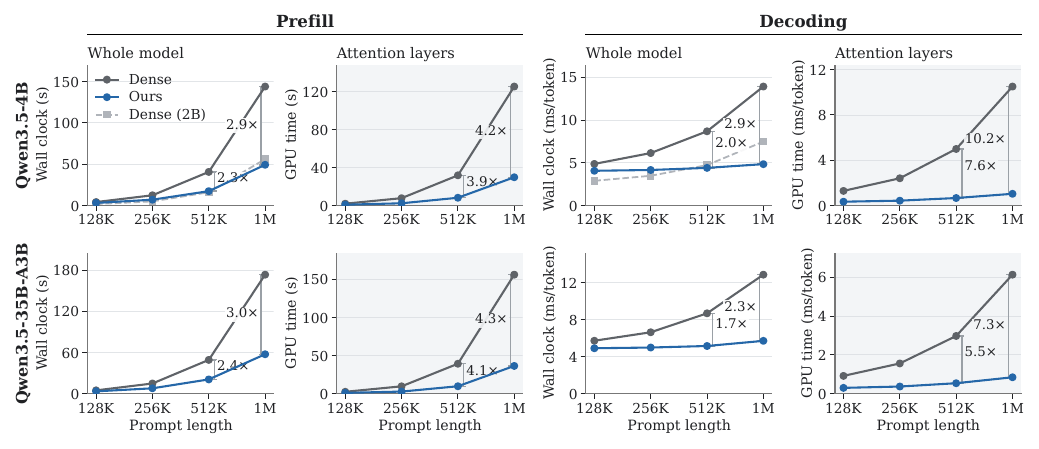}
\caption{Prefill and decode latency on one H200 with a token budget of $L/16$ for a prompt of $L$ tokens.
Whole-model latency is measured by wall clock, and full-attention layers (shaded) by GPU time.
The rows are Qwen3.5-4B (top) and 35B-A3B (bottom).
Dense (2B) is Qwen3.5-2B.
Speedups are over dense attention (Tables~\ref{tab:runtime} and~\ref{tab:qwen35b}).}
\label{fig:latency}
\end{figure}

We show that, without training, a sparse prefill in which each query selects its own blocks can still run much faster than dense attention.
Our method clusters the keys of each input prompt into large semantic blocks, groups of tokens with similar keys regardless of their positions in the sequence.
Each query scores the cluster centroids and attends to the original keys and values of the blocks it selects.
Unlike contiguous blocks, a semantic block gathers tokens that a query is likely to need together.
We use the same index for both the prompt's own prefill and decoding.

Scoring one centroid per block makes selection cheaper than scoring every key individually.
During prefill, queries that select the same block are processed together for that block, as in the KV-outer execution of MoBA and MSA~\citep{lu2025moba,lai2026msa}.
This lets the GPU reuse KV data in on-chip memory across queries while preserving each query's own block selection.
Larger blocks make centroid scoring and block-wise attention cheaper but token selection less precise.
The time saved leaves room for a larger token budget, which can compensate for the coarser selection (Section~\ref{sec:results-budget}).

We evaluate on hybrid models, which use full attention in only some layers and are increasingly common among top-tier LLMs~\citep{minimax2025m1,qwen2026qwen35,gemma2026gemma4}.
On Qwen3.5-4B and 35B-A3B, whole-model prefill is $2.3$--$2.4\times$ faster and decoding $1.7$--$2.0\times$ faster than dense attention at 512K tokens, and prefill is $2.9$--$3.0\times$ faster at 1M, with index construction included and dense attention run on the same inference stack (Figure~\ref{fig:latency}).
The \mbox{35B-A3B} model stays close to dense attention on five RULER tasks~\citep{hsieh2024ruler} through 1M, on LongBench-v2~\citep{bai2025longbenchv2} overall, and on MRCR~\citep{openai2025mrcr} through 256K.
The 4B model, which loses accuracy mainly on the multi-key tasks in RULER, still outperforms dense Qwen3.5-2B, whose latency is similar at 512K (Table~\ref{tab:quality-main}).

\Needspace{8\baselineskip}
\section{Related work}
\label{sec:related-work}
Many sparse attention methods first score individual tokens or blocks of consecutive tokens, then compute attention over the selected keys and values.
DSA~\citep{deepseek2025v32} and MSA~\citep{lai2026msa} use learned indexers to score every visible token, with MSA pooling the resulting scores to select blocks.
QSA reduces the number of candidates by pooling indexer keys into 4-token blocks before scoring~\citep{qwen2026qsa}.
NSA selects blocks by the attention that queries give to compressed block keys~\citep{yuan2025nsa}.
These systems train their selection together with the language model.
DSA, MSA, and QSA perform well with selected-token budgets of 2,048.

\begin{figure}[t]
\centering
\setlength{\abovecaptionskip}{4pt}
\includegraphics[width=0.92\linewidth]{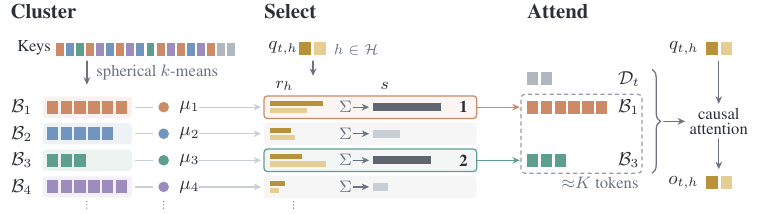}
\caption{Keys are clustered separately for each layer and KV head.
The query heads that share a KV head score the centroids separately ($r_h$), and the sum of their scores ($s$) selects the clusters.
Each head attends to the original KV of the selected clusters, together with recent tokens $\mathcal D_t$, under a causal mask.
Only positions at or before the query contribute to the attention sum.}
\label{fig:concept}
\end{figure}

Quest~\citep{tang2024quest} and MoBA~\citep{lu2025moba} select blocks of consecutive tokens using summaries of the model's existing keys.
MoBA computes attention block by block, together for all queries that selected a block, as MSA also does~\citep{lai2026msa}.
FlashMoBA speeds up this computation for small blocks~\citep{xiao2025flashmoba}.
Training-free prefill methods make selection cheaper with structured patterns and block-level estimates, as in MInference~\citep{jiang2024minference}.
FlexPrefill's query-aware pattern scores blocks using pooled queries and keys~\citep{lai2025flexprefill}, while QuoKA uses representative queries to select a shared KV subset for a chunk~\citep{jones2026quoka}.
FlashPrefill and FlashPrefill V2 score mean-pooled key blocks for each tile of queries and keep the blocks above a threshold relative to the maximum score, the latter compensating the rest with their mean keys and values~\citep{fan2026flashprefill,fan2026flashprefillv2}.
\citet{nawrot2026sparsefrontier} compare training-free methods using FLOPs for prefill and memory traffic for decoding rather than measured latency.
They also note that fine-grained per-query selection during prefill is difficult to accelerate because importance estimation is costly and available kernels struggle with irregular sparsity.

Clustering groups tokens by key similarity rather than sequence position.
Routing Transformer trains a model whose queries and keys are routed by spherical $k$-means~\citep{roy2020routing}.
SAAP fits a key partition on a corpus and trains a query classifier to select clusters while keeping the language model frozen~\citep{mazare2025saap}.
Squeezed Attention~\citep{hooper2025squeezed}, ClusterKV~\citep{liu2025clusterkv}, and Tactic~\citep{zhu2026tactic} use query--centroid scores without a learned classifier and attend to the selected original keys and values.
Squeezed Attention indexes a reusable context offline, whereas ClusterKV and Tactic build prompt-specific clusters during dense prefill for subsequent sparse decoding.
RetrievalAttention similarly searches a nearest-neighbor index of the keys during decoding~\citep{liu2024retrievalattention}.
Our method also builds its clusters for each prompt without training, but uses them for the prompt's own prefill as well as its decoding.
Its clusters are also much larger, 2,048 tokens in the main setting against tens of tokens in these methods, which makes centroid scoring and KV-outer attention cheap enough to afford a larger budget, with a separate selection for each query.

The inference-time methods above, which leave the language model unchanged, are evaluated in their own papers on Transformers in which every layer uses softmax attention (Appendix~\ref{app:related-table}).
UniPrefill, which skips token computation in attention and non-attention layers, measures several of the prefill methods on a linear-attention hybrid up to 128K tokens, where they speed up the time to first token by at most $1.11\times$~\citep{fan2026uniprefill}.
Concurrent work on Qwen3.5 trains a selector and keeps the prefill dense~\citep{ahmad2026spotattention}.
When every layer uses softmax attention, most heads attend locally~\citep{wu2024retrievalhead,xiao2024duoattention}, and turning 85\% of them into sliding-window attention, with decode-time sparse token selection in the rest, loses little accuracy after a few hundred training steps~\citep{zhou2026rtpurbo}.
In hybrids such as MiniMax-M1~\citep{minimax2025m1} or Qwen3.5~\citep{qwen2026qwen35}, a quarter or fewer of the layers use full attention.
These layers provide the model's only exact access to distant tokens, so results on all-attention models do not show how much sparsity they can tolerate.
They also account for only part of the model's cost, so attention-layer speedups alone do not show the whole-model gain.
We therefore report whole-model wall-clock latency and attention-layer GPU time separately, and break down the costs of indexing, selection, and attention execution.

\section{Method}
\label{sec:method}

Our method groups prompt keys into semantic blocks, selects blocks for each query, and computes attention over the selected tokens (Figure~\ref{fig:concept}).
We choose the block size and token budget together to balance task accuracy and whole-model latency, including the cost of building and using the index.
All model weights remain fixed, and no selector is trained.

We describe one full-attention layer and one KV head shared by a group $\mathcal H$ of query heads under grouped-query attention (GQA)~\citep{ainslie2023gqa}.
Let $L$ be the prompt length and $d$ the attention head dimension.
We write $k_i,v_i\in\mathbb R^d$ for the key and value at position $i$, and $q_{t,h}\in\mathbb R^d$ for the query at position $t$ in head $h\in\mathcal H$.
We omit layer and KV-head indices.

\subsection{Building semantic blocks}
\label{sec:prompt-partition}

Let $\tilde k_i$ and $\tilde q_{t,h}$ denote the keys and queries normalized to unit $L_2$ norm.
For each prompt, layer, and KV head, we run spherical $k$-means on a sample of the normalized keys to obtain $C$ unit-norm centroids $\mu_1,\ldots,\mu_C$.
We assign each prompt key to its nearest centroid and define its block as
\begin{equation}
 a_i=\operatorname*{arg\,max}_{c\in\{1,\ldots,C\}}\tilde k_i^\top\mu_c,
 \qquad
 \mathcal B_c=\{i\in\{1,\ldots,L\}:a_i=c\}.
 \label{eq:semantic-blocks}
\end{equation}

\begin{wrapfigure}[15]{r}{0.40\textwidth}
\vspace{-12pt}
\centering
\includegraphics[width=\linewidth]{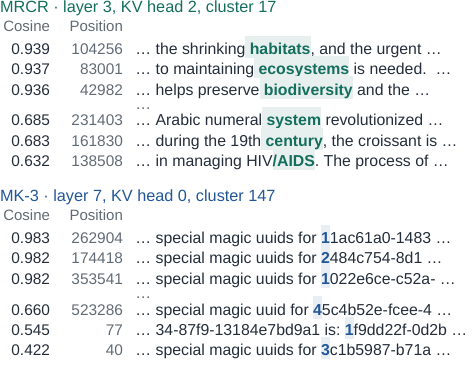}
\caption{Examples of two clusters, with members of high and low cosine similarity to the centroid.}
\label{fig:cluster-examples}
\vspace{-8pt}
\end{wrapfigure}
Each $\mathcal B_c$ contains the positions of tokens assigned to cluster $c$.
We call these groups \emph{semantic blocks} because they group similar keys regardless of their positions in the sequence.
Figure~\ref{fig:cluster-examples} and Appendix~\ref{app:cluster-examples} show tokens that the clustering places in the same block.
A block can contain many different tokens.
In the upper example, the members closest to the centroid are words about the environment from distant passages.
In the lower example, they are the leading digits of different identifiers.

A target mean block size $u$ determines the cluster count approximately as $C=L/u$ (Appendix~\ref{app:implementation} gives the clustering details).
Each layer and KV head has its own partition, and the members of a block are generally scattered across the prompt, so we store each block's keys and values contiguously (Section~\ref{sec:implementation}).

\subsection{Selecting blocks and computing attention}
\label{sec:gqa-routing}

Prefill queries are processed in contiguous chunks.
Each query attends to all causally visible tokens in its current chunk and selects clusters from the earlier history.
During decoding, recent generated tokens are similarly included without selection.
Let $\mathcal D_t\subseteq\{1,\ldots,t\}$ denote the positions included without selection for the query at position $t$.
The remaining visible positions in cluster $c$ are $\mathcal B'_{c,t}=(\mathcal B_c\cap\{1,\ldots,t\})\setminus\mathcal D_t$.

We compute softmax scores over centroids within each query head and sum them across the GQA group before selecting blocks.
We also apply a size penalty to account for the number of tokens retrieved from each cluster,
\begin{equation}
 r_h(c)=\frac{\exp(\tilde q_{t,h}^\top\mu_c/\tau)}{\sum_{c'=1}^{C}\exp(\tilde q_{t,h}^\top\mu_{c'}/\tau)},
 \qquad
 s(c)=\frac{\sum_{h\in\mathcal H}r_h(c)}{(1+|\mathcal B'_{c,t}|)^\beta}.
 \label{eq:gqa-score}
\end{equation}
We use temperature $\tau=0.1$ and size-penalty exponent $\beta=0.25$.
We select clusters in decreasing order of $s(c)$, taking all tokens in each $\mathcal B'_{c,t}$ until their total number reaches or exceeds the budget $K$.
Let $\mathcal J_t$ be the clusters selected at position $t$.
The positions of the key--value pairs used for attention are
\begin{equation}
 \mathcal S_t=\mathcal D_t\;\cup\!\bigcup_{c\in\mathcal J_t}\mathcal B'_{c,t}.
 \label{eq:budget-selection}
\end{equation}
This selection is computed for each query position and shared by the query heads in its GQA group.

Attention is then computed over these positions using the model's original queries, keys, and values\footnote{For models with RoPE~\citep{su2024roformer} or partial RoPE, our experiments use queries and keys before the rotation for clustering and selection. Using them after the rotation gives similar accuracy (Appendix~\ref{app:routing-space}).},
\begin{equation}
 o_{t,h}=
 \frac{\sum_{i\in\mathcal S_t}\exp(q_{t,h}^\top k_i/\sqrt{d})\,v_i}
 {\sum_{i\in\mathcal S_t}\exp(q_{t,h}^\top k_i/\sqrt{d})}.
 \label{eq:sparse-attention}
\end{equation}
Setting $\mathcal S_t=\{1,\ldots,t\}$ recovers dense attention.

\subsection{Prefill and decoding}
\label{sec:prefill-decode}

During prefill, we build the index from the prompt keys at each layer and process the prompt queries in contiguous chunks.
Each query attends to selected blocks from earlier chunks and to all causally visible tokens in its own chunk.
For queries near the beginning of the prompt, the earlier history is short, so we avoid selection overhead by using dense attention over the whole visible prefix, setting $\mathcal D_t=\{1,\ldots,t\}$ (see Appendix~\ref{app:implementation}).

During decoding, each query uses the same index to select prompt blocks.
In our main experiments, all generated tokens are included without selection in $\mathcal D_t$.
For longer generations, older generated tokens can instead be added to the index, leaving only recent ones in $\mathcal D_t$ (Section~\ref{sec:results-generation}).

For simplicity, our main experiments run spherical $k$-means on keys sampled from the whole prompt.
Block selection for a prompt query can therefore depend on later prompt keys, although each query attends only to positions at or before it.
This is possible because the whole prompt is given before generation starts, and every generated token still depends only on the tokens before it.
Because the index is fitted to keys drawn from the whole prompt, prefill runs one layer at a time over the whole prompt, instead of passing chunks of the prompt through all layers as ordinary chunked prefill does (Appendices~\ref{app:prefill-schedules} and~\ref{app:memory}).
Alternatively, running $k$-means only on the first 64K keys and assigning each later key to its nearest centroid allows ordinary chunked prefill, with similar accuracy (Section~\ref{sec:results-selection} and Appendix~\ref{app:memory}).

\subsection{Efficient execution}
\label{sec:implementation}
Efficient dense-attention kernels such as FlashAttention-3~\citep{shah2024flash3} process neighboring queries together so that they can reuse each KV tile.
In our method, however, neighboring queries can select different KV blocks.
If they are still processed together, a block selected by one query may also be processed for queries that did not select it, reducing the savings from sparsity.
We instead gather the queries that selected each KV block and compute their attention to that block together.
MoBA and MSA use the same execution strategy for blocks of consecutive tokens~\citep{lu2025moba,lai2026msa}.

For each prefill chunk and KV head, FlashAttention-3 processes a batch of KV blocks paired with their query groups.
Attention within the current chunk uses causal FlashAttention-3.
For each query, we combine the attention outputs using their softmax normalizers (Appendix~\ref{app:implementation}).
Following MSA~\citep{lai2026msa}, we call this block-wise execution \emph{KV-outer}.
On Qwen3.5-4B at 512K, prefill with KV-outer is $2.6\times$ faster than processing neighboring queries over their selected-block union (Appendix~\ref{app:ablation}).

Although a block's tokens may be scattered across the prompt, we store their KV contiguously in their original sequence order.
For each prefill chunk, we can then read the KV of earlier tokens directly from the start of each block.
We build this block-ordered cache once per layer and use it for both prefill and decoding.
Without this layout, gathering the selected KV for every prefill chunk increases 512K prefill time by about 9\% (Appendix~\ref{app:ablation}).

During decoding, each step reads prompt KV from the selected blocks rather than the full prompt.
This reduces the prompt tokens read per step from $L$ to roughly $K$.
Our vLLM implementation~\citep{kwon2023vllm} uses fused kernels to reduce the overhead of scoring and selecting blocks during decoding.

Each query scores $C$ centroids instead of the prompt's $L$ keys.
For a fixed prompt and token budget, smaller blocks give finer selection but require scoring more centroids and processing more blocks~\citep{xiao2025flashmoba}.
Larger blocks reduce this overhead, so we can increase the budget to compensate for coarser selection (Section~\ref{sec:results-budget}).
In our experiments, both $C$ and $K$ grow in proportion to $L$, so prefill remains quadratic in prompt length, but with a smaller constant factor than dense attention.
Appendix~\ref{app:complexity} gives the cost of each step.

\section{Experimental setup}
\label{sec:experiments}

\paragraph{Models and baselines.}
We evaluate Qwen3.5-4B and Qwen3.5-35B-A3B~\citep{qwen2026qwen35}, replacing their full-attention layers (8 of 32 and 10 of 40) while keeping all weights fixed, and also apply the method to a hybrid of another family, Nemotron 3.5 Lightning~\citep{nvidia2026nemotron35} (Appendix~\ref{app:nemotron-section}).
Dense attention runs on the same vLLM build with chunked prefill, and dense Qwen3.5-2B serves as a smaller-model reference.
As a controlled sparse baseline, we replace semantic blocks with positional blocks of $B$ consecutive tokens, each represented by its mean key.
As in block-sparse prefill methods~\citep{jiang2024minference,lai2025flexprefill,fan2026flashprefillv2}, $g$ neighboring queries can share one block selection to reduce its cost.
It uses the same token budget, the same $\mathcal D_t$, and the same attention kernels as our method.
This gives a controlled comparison with positional block selection and query sharing.

\paragraph{Benchmarks.}
We use RULER~\citep{hsieh2024ruler}, eight-needle MRCR~\citep{openai2025mrcr}, and LongBench-v2~\citep{bai2025longbenchv2}.
RULER-5 averages five RULER tasks, namely multi-key retrieval (MK-2 and MK-3), multi-value retrieval, common-word extraction, and HotpotQA.
Most other tasks are too easy to separate the methods at 512K, and the 4B model's average over all thirteen stays close to dense attention (Appendix~\ref{app:ruler-by-task}).
RULER and MRCR use about 100 examples per task and length, and LongBench-v2 uses the 470 of 503 examples that fit without truncation.
RULER uses greedy decoding without thinking, while MRCR and LongBench-v2 use thinking with Qwen's recommended sampling settings.
We average our main results on all three benchmarks over three runs (Appendix~\ref{app:protocol}).

\paragraph{Settings.}
Both Qwen models use the same configuration, with target mean block size $u=2{,}048$ and budget $K=L/16$ for RULER and latency measurements.
Most ablations use the 4B model.
MRCR and LongBench-v2 have variable-length prompts, so we set budgets by length range.
For MRCR, $K$ is one sixteenth of the bin's upper length bound.
For LongBench-v2, $K=16{,}384$ for prompts up to 256K and $K=65{,}536$ for longer prompts.
The budget excludes the tokens attended without selection and may be exceeded by the last selected block.
Beyond the native 256K context, dense and sparse runs both use factor-4 YaRN interpolation~\citep{peng2024yarn} without its attention scaling.
This setting gave the best dense-attention accuracy in our tests (Appendix~\ref{app:positional-scaling}).

\section{Results}
\label{sec:results}

\subsection{Overall performance}
\label{sec:results-latency}
\label{sec:results-quality}

\begin{table}[t]
\centering
\small
\caption{Long-context performance at $K=L/16$, with $L$ the upper bound of the length bin for MRCR-8 and of the length band for LongBench-v2, and whole-model prefill latency at 1M.
MRCR-8 reports a match ratio, and the other quality scores are percentages.
At 1M, the multi-value prompts of RULER-5 are about 647K tokens long (Appendix~\ref{app:prompt-lengths}).
Appendix~\ref{app:budget-scaling} gives other budgets.}
\label{tab:quality-main}
\setlength{\tabcolsep}{4.2pt}
\begin{tabular}{l cccccccccc}
\toprule
 & \multicolumn{3}{c}{RULER-5} & \multicolumn{3}{c}{MRCR-8} & \multicolumn{3}{c}{LongBench-v2} & Prefill (s) \\
\cmidrule(lr){2-4} \cmidrule(lr){5-7} \cmidrule(lr){8-10} \cmidrule(lr){11-11}
 & 256K & 512K & 1M & 128K & 256K & 512K & $\le$256K & $>$256K & all & 1M \\
\midrule
35B-A3B, dense & 85.7 & 84.4 & 73.2 & 0.846 & 0.655 & 0.564 & 64.4 & 66.2 & 64.7 & 173.5 \\
35B-A3B, ours & 84.5 & 84.2 & 73.9 & 0.844 & 0.668 & 0.504 & 63.1 & 64.7 & 63.3 & 57.4 \\
\midrule
4B, dense & 84.7 & 82.8 & 63.2 & 0.797 & 0.574 & 0.412 & 54.3 & 64.2 & 55.7 & 144.1 \\
4B, ours & 82.7 & 81.8 & 55.2 & 0.788 & 0.561 & 0.399 & 53.9 & 57.7 & 54.5 & 49.3 \\
\midrule
\textcolor{black!55}{2B, dense} & \textcolor{black!55}{77.7} & \textcolor{black!55}{66.5} & \textcolor{black!55}{48.9} & \textcolor{black!55}{0.726} & \textcolor{black!55}{0.414} & \textcolor{black!55}{0.198} & \textcolor{black!55}{36.7} & \textcolor{black!55}{34.3} & \textcolor{black!55}{36.4} & \textcolor{black!55}{56.2} \\
\bottomrule
\end{tabular}

\end{table}

We measure latency at batch size one on a single H200 in BF16, with FlashAttention-3 for prefill and full-step CUDA graphs for decoding in both methods, and report the median of three repetitions after warm-up (Appendix~\ref{app:protocol}).
Prefill latency includes index construction, KV packing, selection, gathering, and merging.
On Qwen3.5-4B and 35B-A3B, whole-model prefill is $2.3$--$2.4\times$ faster and decoding $1.7$--$2.0\times$ faster than with dense attention at 512K (Figure~\ref{fig:latency}).
The gains grow with context length, reaching $2.9$--$3.0\times$ for prefill and $2.3$--$2.9\times$ for decoding at 1M.
The full-attention layers themselves prefill $3.9$--$4.3\times$ faster at these lengths, and the whole-model speedup is lower because the other layers are unchanged (Tables~\ref{tab:runtime} and~\ref{tab:qwen35b}).

\begin{wrapfigure}{r}{0.49\textwidth}
\vspace{-12pt}
\centering
\includegraphics[width=\linewidth]{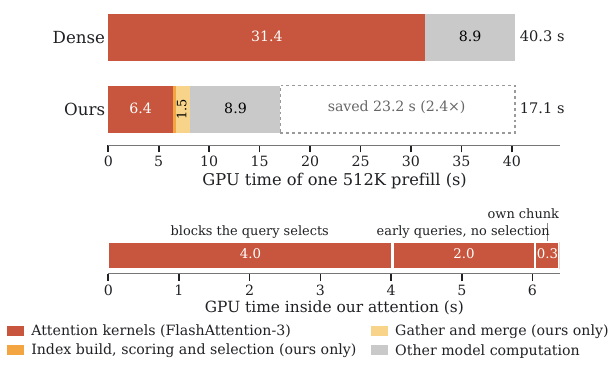}
\caption{GPU-time breakdown for a 512K prefill on Qwen3.5-4B.
The upper panel shows whole-model GPU time, and the lower panel breaks down our attention-kernel time (Table~\ref{tab:stage-breakdown}).
Times are profiler totals, not wall-clock latency.}
\label{fig:breakdown}
\vspace{-8pt}
\end{wrapfigure}

Figure~\ref{fig:breakdown} breaks down prefill time on Qwen3.5-4B at 512K.
Our attention kernels run about $5\times$ faster than dense attention.
Index construction, selection, gathering, and merging together take about 1.8\,s, of which fitting the centroids is a small part, compared with roughly 25\,s saved in attention computation.
The rest of the model is unchanged and takes about half of the sparse prefill time.
We use dense attention within query chunks and near the start of the prompt, so an $L/16$ budget does not give a $16\times$ attention speedup.
We also tried switching from dense to sparse attention earlier in the prompt, but this did not speed up prefill.
The extra selection, gathering, and merging offset the time saved on dense attention (Appendices~\ref{app:pipeline-stages} and~\ref{app:dense-prefix}).

Both models retain most of their dense-attention performance across the three benchmarks (Table~\ref{tab:quality-main}).
On the 35B-A3B model, RULER-5 remains close to dense attention through 1M.
The 4B model loses more on multi-key retrieval, especially at 1M, while multi-value retrieval and HotpotQA remain close to dense attention across lengths (Appendix~\ref{app:ruler-by-task}).
MRCR-8 scores stay close to dense attention through 256K on both models and fall below it at 512K, more on the 35B-A3B model.
Overall LongBench-v2 accuracy changes little on either model.
A smaller dense model also reduces latency, but the sparse 4B model scores higher than dense Qwen3.5-2B on RULER-5 at every tested length, on MRCR-8, and on LongBench-v2, with similar prefill latency at 512K and lower latency at 1M (Table~\ref{tab:quality-main} and Appendix~\ref{app:latency-4b}).
On Nemotron 3.5 Lightning, with its own clustering settings, multi-key accuracy falls further at $L/16$, so we use $K=L/8$.
Whole-model prefill is then $1.5$--$2.0\times$ faster than dense attention over prompts from 256K to 1M.
At 512K, accuracy remains close to dense attention except on multi-key retrieval (Appendix~\ref{app:nemotron-section}).

\subsection{Semantic blocks and per-query selection}
\label{sec:results-positional}

\begin{figure}[b]
\centering
\setlength{\abovecaptionskip}{4pt}
\includegraphics[width=\linewidth,trim=0 9 0 5,clip]{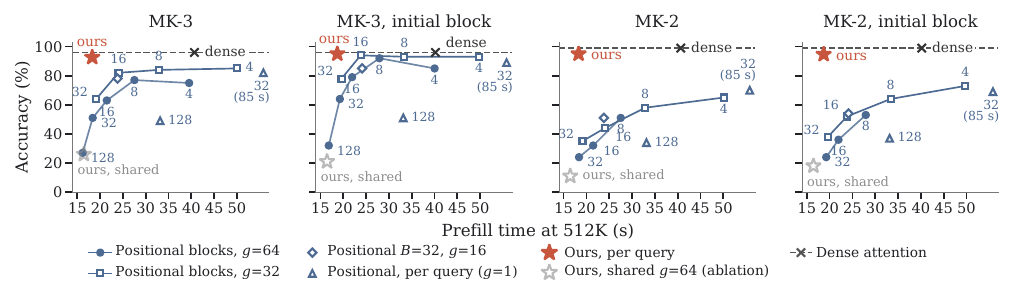}
\caption{MK-3 and MK-2 accuracy against prefill latency on Qwen3.5-4B at 512K, without and with an initial block of 1,024 positions.
Numbers are the positional block size $B$, and $g$ is the number of queries sharing one selection.
Adding the initial block has little effect on prefill time (Appendix~\ref{app:positional}).
We therefore reuse timings measured without it for per-query selection and shared semantic selection.}
\label{fig:positional}
\end{figure}

We compare our method with positional block selection using the baseline in Section~\ref{sec:experiments}.
We vary the positional block size $B$ and the number $g$ of neighboring queries sharing a selection.
For shared selection, we combine scores from all queries in the group to choose one set of blocks, following FlashPrefill V2~\citep{fan2026flashprefillv2}.
Sharing reduces cost but can miss blocks needed by individual queries.
We also tried averaging the queries before scoring, or taking the maximum query score for each block.
Neither improved accuracy (Appendix~\ref{app:positional}).

Figure~\ref{fig:positional} compares these choices at the same token budget on multi-key retrieval, where the 4B model is most sensitive to sparsification.
Large positional blocks are fast but retrieve poorly, and smaller ones improve accuracy at a higher execution cost, particularly with per-query selection.
At $B=32$, per-query positional selection is still less accurate than our method and more than $4\times$ slower to prefill.
Sharing selections makes positional blocks faster, but the most accurate shared settings remain less accurate than our method and take $1.8$--$2.7\times$ as long to prefill.
On MK-2 the gap is wider, with every positional setting at least 25 points below our method.
On Qwen3.5-35B-A3B, only one tested positional setting matches our MK-3 accuracy, but it takes $1.8\times$ as long to prefill and scores 14 points lower on MK-2 (Appendix~\ref{app:positional}).
As a reference point, we also ran FlashPrefill V2 with the SGLang implementation released with it, on the same prompts (Appendix~\ref{app:flashprefill}).
The two engines differ in implementation, but their dense attention runs at a similar speed and reaches a similar accuracy.
We compare speedups relative to dense attention on each method's own engine.
Our method is more accurate at comparable speedups.
Among the tested thresholds, FlashPrefill V2 reaches our MK-3 accuracy only with a $1.7\times$ speedup, compared with $2.2\times$ for ours.

We also test shared selection for semantic blocks, using groups of 64 neighboring queries as in MInference's block-sparse pattern~\citep{jiang2024minference}.
This sharply reduces MK-3 accuracy for only a small reduction in prefill time.
Per-query selection of semantic blocks thus gives the best accuracy--latency trade-off among these configurations.
At a fixed token budget, larger blocks mean fewer blocks per query, reducing the cost of gathering queries and merging outputs (Section~\ref{sec:implementation}).
At 512K, each query scores 256 semantic centroids, compared with 16,384 block means for 32-token positional blocks.

We also test adding a fixed initial block of 1,024 positions, following the practice of including early prompt positions to retain attention sinks~\citep{xiao2024efficient,jiang2024minference}.
This tends to improve both methods. Ours then matches the most accurate positional setting on MK-3 while prefilling $1.3\times$ faster, and remains far more accurate on MK-2, also at the smaller budget $K=L/32$ (Figure~\ref{fig:positional}, Table~\ref{tab:positional-points} and Appendix~\ref{app:positional}).

\subsection{Block size and token budget}
\label{sec:results-budget}
\label{sec:results-breakdown}
\label{sec:results-systems}
\label{sec:results-transfer}

\begin{figure}[t]
\begin{minipage}[t]{0.48\linewidth}
\centering
\includegraphics[width=\linewidth]{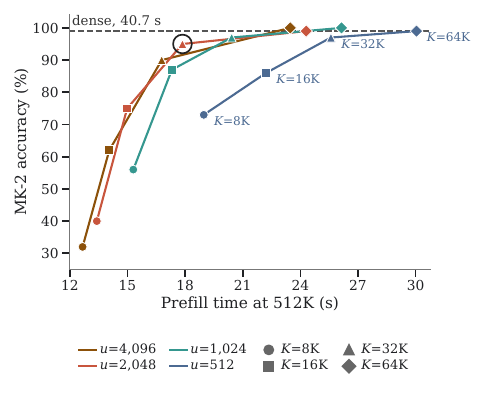}
\caption{Cluster size and token budget on MK-2 at 512K, Qwen3.5-4B.
Larger clusters with a larger budget can improve both accuracy and latency over finer clusters with a smaller budget.}
\label{fig:operating-point}
\end{minipage}\hfill
\begin{minipage}[t]{0.50\linewidth}
\centering
\includegraphics[width=\linewidth]{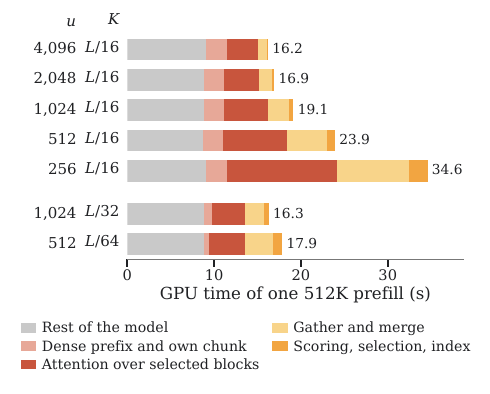}
\caption{GPU time of one 512K prefill on Qwen3.5-4B by cluster size $u$ and token budget $K$.}
\label{fig:cluster-size-profile}
\end{minipage}
\end{figure}

On Qwen3.5-4B at 512K, smaller clusters are more accurate at small token budgets.
As the budget grows, larger clusters reach similar accuracy while remaining faster (Figure~\ref{fig:operating-point}; Appendix~\ref{app:operating-point}).
Our main setting uses 2,048-token clusters at $K=L/16$.
We also analyze the attention mass retained by the selected tokens and test selection using dense-attention scores (Appendix~\ref{app:mass-decomposition}).

Figure~\ref{fig:cluster-size-profile} breaks down prefill GPU time by cluster size.
At a fixed budget, reducing the cluster size from 2,048 to 512 tokens increases prefill GPU time by 41\% on the 4B model.
Most of the increase comes from attention over selected blocks, query gathering, and output merging.
With 2,048-token clusters, attention over selected blocks reaches about 70\% of the FLOP rate of dense FlashAttention-3 on the same GPU, compared with less than 40\% with 512-token clusters.
Kernels specialized for small blocks~\citep{xiao2025flashmoba} could narrow this gap.

On Nemotron, smaller clusters give larger accuracy gains with a smaller runtime penalty.
Reducing the cluster size from 2,048 to 512 tokens increases prefill GPU time by 27\% (Figure~\ref{fig:nemotron-size-profile}).
Clusters around 512 to 1,024 tokens work well on this model.
Cluster size and token budget should therefore be chosen together for each model and its kernels.

\subsection{Prompt-specific clustering}
\label{sec:results-selection}
\label{sec:results-coverage}

\begin{wraptable}{r}{0.47\textwidth}
\vspace{-10pt}
\centering
\small
\caption{Centroid fitting data on Qwen3.5-4B at 512K ($K=L/16$).}
\label{tab:centroid-source}
\setlength{\tabcolsep}{2.5pt}
\resizebox{\linewidth}{!}{\begin{tabular}{lcccc}
\toprule
Centroids fitted on & MK-3 & MK-2 & Hotpot & CWE \\
\midrule
Dense attention & 96.0 & 99.0 & 56.0 & 63.0 \\
Each prompt (ours) & 92.7 & 95.0 & 57.0 & 64.3 \\
Each prompt's first 64K & 95.0 & 95.0 & 56.0 & 66.1 \\
A raw-text corpus & 1.0 & 12.0 & 51.0 & \textit{failed} \\
RULER-5 examples & 91.0 & 90.0 & 55.0 & 64.6 \\
\bottomrule
\end{tabular}
}
\vspace{-8pt}
\end{wraptable}

Fitting centroids once on a shared corpus allows reuse across prompts, as in SAAP~\citep{mazare2025saap}.
We compare this approach with our prompt-specific fitting, keeping the selection rule unchanged (Table~\ref{tab:centroid-source}).
Centroids fitted on a raw-text corpus lose almost all multi-key retrieval accuracy.
Using RULER-5 evaluation prompts as the fitting corpus largely restores it, and fitting each prompt separately gives a further gain on MK-2 (Appendix~\ref{app:routing-variants}).

Fitting centroids on just the first 64K tokens of each prompt gives similar accuracy to using the whole prompt and allows ordinary chunked prefill with less memory (Section~\ref{sec:prefill-decode} and Appendix~\ref{app:memory}).

\subsection{Decoding over long generations}
\label{sec:results-generation}

In our main experiments, each decoding step attends to all generated tokens without selection.
Decoding therefore slows as the output grows, which matters for the long reasoning traces in MRCR-8 and LongBench-v2 (blue in Figure~\ref{fig:generated-length}; Appendix~\ref{app:suffix-flush}).

We can instead add older generated tokens to the index by assigning them to the existing centroids without refitting.
We update the index every 1,024 generated tokens and keep the newest 512 tokens outside it.
Decode latency then stays nearly constant (orange in Figure~\ref{fig:generated-length}), even over 64K generated tokens after a 1M-token prompt.
LongBench-v2 accuracy changes little on either Qwen model (Appendices~\ref{app:suffix-flush} and~\ref{app:qwen35b}).

\section{Limitations and future work}
\label{sec:limitations}

Our experiments use BF16 inference at batch size one on a single H200.
Throughput under continuous batching and on other GPU architectures remains to be evaluated.
The method retains the full prompt KV cache, so memory capacity still limits context length and batch size.
Fitting centroids on the whole prompt raises peak memory above dense chunked prefill, but fitting them on the first 64K tokens allows chunked prefill with similar accuracy (Appendix~\ref{app:memory}).
Our method can also be combined with other efficiency techniques, such as KV-cache quantization~\citep{liu2024kivi} and cache compression~\citep{li2024snapkv}, which reduce different costs.

Training the model for sparse attention, as in DSA~\citep{deepseek2025v32} and QSA~\citep{qwen2026qsa}, may bring further gains.
Our short distillation experiments did not consistently improve retrieval (Appendix~\ref{app:loss}), but longer training on long-context task data may help preserve accuracy at smaller budgets.

\section{Conclusion}
\label{sec:conclusion}

We presented a simple but effective training-free sparse attention method that builds semantic blocks from each prompt and uses them for both prefill and decoding.
Centroid scoring keeps selection cheap, while KV-outer execution makes prefill efficient with a separate selection for each query.
This addresses the selection and execution costs discussed by \citet{nawrot2026sparsefrontier}.
The resulting savings allow a larger token budget and yield whole-model speedups on two hybrid LLMs.

\subsubsection*{Acknowledgments}
We thank the National Institute of Information and Communications Technology (NICT) for the opportunity to conduct this research under our joint research framework.

\bibliography{references}

@misc{qwen2026qwen35,
  author        = {{Qwen Team}},
  title         = {{Qwen3.5}},
  year          = {2026},
  howpublished  = {\url{https://huggingface.co/collections/Qwen/qwen35}}
}

@misc{gemma2026gemma4,
  author        = {{Gemma Team}},
  title         = {{Gemma 4}},
  year          = {2026},
  url           = {https://huggingface.co/collections/google/gemma-4}
}

@article{mazare2025saap,
  author        = {Pierre-Emmanuel Mazar{\'e} and Gergely Szilvasy and Maria Lomeli and Francisco Massa and Naila Murray and Herv{\'e} J{\'e}gou and Matthijs Douze},
  title         = {Inference-Time Sparse Attention with Asymmetric Indexing},
  journal       = {arXiv preprint \href{https://arxiv.org/abs/2502.08246}{arXiv:2502.08246}},
  year          = {2025}
}

@inproceedings{tang2024quest,
  author        = {Jiaming Tang and Yilong Zhao and Kan Zhu and Guangxuan Xiao and Baris Kasikci and Song Han},
  title         = {{Quest}: Query-Aware Sparsity for Efficient Long-Context {LLM} Inference},
  booktitle     = {International Conference on Machine Learning},
  year          = {2024},
  url           = {https://proceedings.mlr.press/v235/tang24l.html}
}

@inproceedings{liu2024retrievalattention,
  author        = {Di Liu and Meng Chen and Baotong Lu and Huiqiang Jiang and Zhenhua Han and Qianxi Zhang and Qi Chen and Chengruidong Zhang and Bailu Ding and Kai Zhang and Chen Chen and Fan Yang and Yuqing Yang and Lili Qiu},
  title         = {{RetrievalAttention}: Accelerating Long-Context {LLM} Inference via Vector Retrieval},
  booktitle     = {Advances in Neural Information Processing Systems},
  volume        = {38},
  year          = {2025},
  note          = {\href{https://arxiv.org/abs/2409.10516}{arXiv:2409.10516}}
}

@inproceedings{jiang2024minference,
  author        = {Huiqiang Jiang and Yucheng Li and Chengruidong Zhang and Qianhui Wu and Xufang Luo and Surin Ahn and Zhenhua Han and Amir H. Abdi and Dongsheng Li and Chin-Yew Lin and Yuqing Yang and Lili Qiu},
  title         = {{MInference} 1.0: Accelerating Pre-filling for Long-Context {LLM}s via Dynamic Sparse Attention},
  booktitle     = {Advances in Neural Information Processing Systems},
  year          = {2024},
  url           = {https://proceedings.neurips.cc/paper_files/paper/2024/hash/5dfbe6f5671e82c76841ba687a8a9ecb-Abstract-Conference.html}
}

@inproceedings{yuan2025nsa,
  author        = {Jingyang Yuan and Huazuo Gao and Damai Dai and Junyu Luo and Liang Zhao and Zhengyan Zhang and Zhenda Xie and Y. X. Wei and Lean Wang and Zhiping Xiao and Yuqing Wang and Chong Ruan and Ming Zhang and Wenfeng Liang and Wangding Zeng},
  title         = {Native Sparse Attention: Hardware-Aligned and Natively Trainable Sparse Attention},
  booktitle     = {Proceedings of the 63rd Annual Meeting of the Association for Computational Linguistics},
  year          = {2025},
  note          = {\href{https://arxiv.org/abs/2502.11089}{arXiv:2502.11089}}
}

@article{deepseek2025v32,
  author        = {{DeepSeek-AI}},
  title         = {{DeepSeek-V3.2}: Pushing the Frontier of Open Large Language Models},
  journal       = {arXiv preprint \href{https://arxiv.org/abs/2512.02556}{arXiv:2512.02556}},
  year          = {2025}
}

@article{lai2026msa,
  author        = {Xunhao Lai and Weiqi Xu and Yufeng Yang and Qiaorui Chen and Yang Xu and Lunbin Zeng and Xiaolong Li and Haohai Sun and Haichao Zhu and Vito Zhang and Jinkai Hu and Jiayao Li and Rui Gao and Zekun Li and Songquan Zhu and Jingkai Zhou and Pengyu Zhao},
  title         = {{MiniMax Sparse Attention}},
  journal       = {arXiv preprint \href{https://arxiv.org/abs/2606.13392}{arXiv:2606.13392}},
  year          = {2026}
}

@misc{openai2025mrcr,
  author        = {{OpenAI}},
  title         = {{OpenAI MRCR}: Long Context Multiple Needle in a Haystack Benchmark},
  year          = {2025},
  howpublished  = {Hugging Face dataset, \url{https://huggingface.co/datasets/openai/mrcr}}
}

@article{roy2020routing,
  author        = {Aurko Roy and Mohammad Saffar and Ashish Vaswani and David Grangier},
  title         = {Efficient Content-Based Sparse Attention with Routing Transformers},
  journal       = {Transactions of the Association for Computational Linguistics},
  volume        = {9},
  year          = {2021},
  url           = {https://aclanthology.org/2021.tacl-1.4/}
}

@inproceedings{kwon2023vllm,
  author        = {Woosuk Kwon and Zhuohan Li and Siyuan Zhuang and Ying Sheng and Lianmin Zheng and Cody Hao Yu and Joseph E. Gonzalez and Hao Zhang and Ion Stoica},
  title         = {Efficient Memory Management for Large Language Model Serving with {PagedAttention}},
  booktitle     = {Proceedings of the 29th ACM Symposium on Operating Systems Principles},
  year          = {2023},
  url           = {https://doi.org/10.1145/3600006.3613165}
}

@inproceedings{shah2024flash3,
  author        = {Jay Shah and Ganesh Bikshandi and Ying Zhang and Vijay Thakkar and Pradeep Ramani and Tri Dao},
  title         = {{FlashAttention-3}: Fast and Accurate Attention with Asynchrony and Low-Precision},
  booktitle     = {Advances in Neural Information Processing Systems},
  volume        = {37},
  year          = {2024},
  note          = {\href{https://arxiv.org/abs/2407.08608}{arXiv:2407.08608}}
}

@inproceedings{hsieh2024ruler,
  author        = {Cheng-Ping Hsieh and Simeng Sun and Samuel Kriman and Shantanu Acharya and Dima Rekesh and Fei Jia and Yang Zhang and Boris Ginsburg},
  title         = {{RULER}: What's the Real Context Size of Your Long-Context Language Models?},
  booktitle     = {Conference on Language Modeling},
  year          = {2024},
  url           = {https://openreview.net/forum?id=kIoBbc76Sy}
}

@inproceedings{ainslie2023gqa,
  author        = {Joshua Ainslie and James Lee-Thorp and Michiel de Jong and Yury Zemlyanskiy and Federico Lebr{\'o}n and Sumit Sanghai},
  title         = {{GQA}: Training Generalized Multi-Query Transformer Models from Multi-Head Checkpoints},
  booktitle     = {Proceedings of the 2023 Conference on Empirical Methods in Natural Language Processing},
  year          = {2023},
  url           = {https://aclanthology.org/2023.emnlp-main.298/}
}

@article{teamolmo2025olmo3,
  author        = {{Team Olmo}},
  title         = {{Olmo 3}},
  journal       = {arXiv preprint \href{https://arxiv.org/abs/2512.13961}{arXiv:2512.13961}},
  year          = {2025}
}

@article{fan2026flashprefill,
  author        = {Qihang Fan and Huaibo Huang and Zhiying Wu and Juqiu Wang and Bingning Wang and Ran He},
  title         = {{FlashPrefill}: Instantaneous Pattern Discovery and Thresholding for Ultra-Fast Long-Context Prefilling},
  journal       = {arXiv preprint \href{https://arxiv.org/abs/2603.06199}{arXiv:2603.06199}},
  year          = {2026}
}

@inproceedings{xu2026hisa,
  author        = {Yufei Xu and Fanxu Meng and Fan Jiang and Yuxuan Wang and Ruijie Zhou and Zhaohui Wang and Jiexi Wu and Zhixin Pan and Xiaojuan Tang and Wenjie Pei and Tongxuan Liu and Di Yin and Xing Sun and Muhan Zhang},
  title         = {{HISA}: Efficient Hierarchical Indexing for Fine-Grained Sparse Attention},
  booktitle     = {Conference on Language Modeling},
  year          = {2026},
  note          = {\href{https://arxiv.org/abs/2603.28458}{arXiv:2603.28458}}
}

@inproceedings{bai2026indexcache,
  author        = {Yushi Bai and Qian Dong and Ting Jiang and Xin Lv and Zhengxiao Du and Aohan Zeng and Jie Tang and Juanzi Li},
  title         = {{IndexCache}: Accelerating Sparse Attention via Cross-Layer Index Reuse},
  booktitle     = {Conference on Language Modeling},
  year          = {2026},
  note          = {\href{https://arxiv.org/abs/2603.12201}{arXiv:2603.12201}}
}

@article{zhou2026rtpurbo,
  author        = {Yanke Zhou and Yiduo Li and Hanlin Tang and Maohua Li and Kan Liu and Tao Lan and Lin Qu and Yuan Yao and Xiaoxing Ma},
  title         = {Full Attention Strikes Back: Transferring Full Attention into Sparse within Hundred Training Steps},
  journal       = {arXiv preprint \href{https://arxiv.org/abs/2605.16928}{arXiv:2605.16928}},
  year          = {2026}
}

@inproceedings{zhu2026tactic,
  author        = {Kan Zhu and Tian Tang and Qinyu Xu and Zhan Jin and Yile Gu and Zhichen Zeng and Rohan Kadekodi and Liangyu Zhao and Ang Li and Arvind Krishnamurthy and Baris Kasikci},
  title         = {{Tactic}: Adaptive Sparse Attention with Clustering and Distribution Fitting for Long-Context {LLM}s},
  booktitle     = {International Conference on Learning Representations},
  year          = {2026},
  url           = {https://openreview.net/forum?id=tJod11fK1A}
}

@inproceedings{hooper2025squeezed,
  author        = {Coleman Richard Charles Hooper and Sehoon Kim and Hiva Mohammadzadeh and Monishwaran Maheswaran and Sebastian Zhao and June Paik and Michael W. Mahoney and Kurt Keutzer and Amir Gholami},
  title         = {Squeezed Attention: Accelerating Long Context Length {LLM} Inference},
  booktitle     = {Proceedings of the 63rd Annual Meeting of the Association for Computational Linguistics},
  year          = {2025},
  url           = {https://aclanthology.org/2025.acl-long.1568/}
}

@inproceedings{liu2025clusterkv,
  author        = {Guangda Liu and Chengwei Li and Jieru Zhao and Chenqi Zhang and Minyi Guo},
  title         = {{ClusterKV}: Manipulating {LLM} {KV} Cache in Semantic Space for Recallable Compression},
  booktitle     = {Proceedings of the 62nd ACM/IEEE Design Automation Conference},
  year          = {2025},
  url           = {https://doi.org/10.1109/DAC63849.2025.11132479}
}

@article{minimax2025m1,
  author        = {{MiniMax}},
  title         = {{MiniMax-M1}: Scaling Test-Time Compute Efficiently with Lightning Attention},
  journal       = {arXiv preprint \href{https://arxiv.org/abs/2506.13585}{arXiv:2506.13585}},
  year          = {2025}
}

@inproceedings{peng2024yarn,
  title={{YaRN}: Efficient Context Window Extension of Large Language Models},
  author={Peng, Bowen and Quesnelle, Jeffrey and Fan, Honglu and Shippole, Enrico},
  booktitle={International Conference on Learning Representations},
  year={2024},
  url           = {https://openreview.net/forum?id=wHBfxhZu1u}
}

@inproceedings{bai2025longbenchv2,
  title     = {{LongBench} v2: Towards Deeper Understanding and Reasoning on Realistic Long-context Multitasks},
  author    = {Bai, Yushi and Tu, Shangqing and Zhang, Jiajie and Peng, Hao and Wang, Xiaozhi and Lv, Xin and Cao, Shulin and Xu, Jiazheng and Dong, Yuxiao and Tang, Jie and Hou, Lei and Li, Juanzi},
  booktitle = {Proceedings of the 63rd Annual Meeting of the Association for Computational Linguistics},
  year      = {2025},
  url           = {https://aclanthology.org/2025.acl-long.183/}
}

@inproceedings{lu2025moba,
  author        = {Enzhe Lu and Zhejun Jiang and Jingyuan Liu and Yulun Du and Tao Jiang and Chao Hong and Shaowei Liu and Weiran He and Enming Yuan and Yuzhi Wang and Zhiqi Huang and Huan Yuan and Suting Xu and Xinran Xu and Guokun Lai and Yanru Chen and Huabin Zheng and Junjie Yan and Jianlin Su and Yuxin Wu and Yutao Zhang and Zhilin Yang and Xinyu Zhou and Mingxing Zhang and Jiezhong Qiu},
  title         = {{MoBA}: Mixture of Block Attention for Long-Context {LLM}s},
  booktitle     = {Advances in Neural Information Processing Systems},
  volume        = {38},
  year          = {2025},
  note          = {\href{https://arxiv.org/abs/2502.13189}{arXiv:2502.13189}}
}

@article{xiao2025flashmoba,
  author        = {Guangxuan Xiao and Junxian Guo and Kasra Mazaheri and Song Han},
  title         = {Optimizing Mixture of Block Attention},
  journal       = {arXiv preprint \href{https://arxiv.org/abs/2511.11571}{arXiv:2511.11571}},
  year          = {2025}
}

@article{qwen2026qsa,
  author        = {Zihan Qiu and Zekun Wang and Xiao Li and Yanpeng Li and Yang Xu and Yixuan Wang and Huaqing Zhang and Rui Men and Bochao Mao and Chengruidong Zhang and Fan Zhou and Hao Luo and Haofeng Huang and Haoran Lian and Haoyan Huang and Hongqing Chen and Jianwei Zhang and Jing Xu and Junjie Wang and Langshi Chen and Liangyu Wang and Linlang Jiang and Man Yuan and Minmin Sun and Peng Jin and Siqi Zhang and Siyu Wang and Xingzhang Ren and Yakai Wang and Yi Zhang and Yiming Dong and Yizhong Cao and Yubo Ma and Yunfei Mao and Bo Zheng and Dayiheng Liu},
  title         = {On the Design of {Qwen3.8-Next} Architecture: Evaluation, Efficiency, and Training Stability},
  journal       = {arXiv preprint \href{https://arxiv.org/abs/2608.30320}{arXiv:2608.30320}},
  year          = {2026}
}

@inproceedings{lai2025flexprefill,
  author        = {Xunhao Lai and Jianqiao Lu and Yao Luo and Yiyuan Ma and Xun Zhou},
  title         = {{FlexPrefill}: A Context-Aware Sparse Attention Mechanism for Efficient Long-Sequence Inference},
  booktitle     = {International Conference on Learning Representations},
  year          = {2025},
  url           = {https://openreview.net/forum?id=OfjIlbelrT}
}

@inproceedings{yuan2025blasst,
  author        = {Yuan, Jiayi and Shinn, Cameron and Xu, Kai and Cui, Jingze and Klimiashvili, George and Xiao, Guangxuan and Zheng, Perkz and Li, Bo and Zhou, Yuxin and Ye, Zhouhai and You, Weijie and Zheng, Tian and Brown, Dominic and Wang, Pengbo and Hoehnerbach, Markus and Cai, Richard and Demouth, Julien and Owens, John D. and Hu, Xia and Han, Song and Liu, Timmy and Mao, Huizi},
  title         = {{BLASST}: Dynamic {BL}ocked Attention Sparsity via Softmax Thresholding},
  booktitle     = {Proceedings of Machine Learning and Systems},
  volume        = {8},
  year          = {2026},
  url           = {https://proceedings.mlsys.org/paper_files/paper/2026/hash/c6ee784cbe46d854843e4c883a3321ef-Abstract-Conference.html}
}

@inproceedings{le2026sketchwalk,
  author        = {Le, Hoang Anh Duy and Joshi, Sahil and Yang, Zeyu and Xu, Zhaozhuo and Shrivastava, Anshumali},
  title         = {Scout Before You Attend: Sketch-and-Walk Sparse Attention for Efficient {LLM} Inference},
  booktitle     = {International Conference on Machine Learning},
  year          = {2026},
  note          = {\href{https://arxiv.org/abs/2602.07397}{arXiv:2602.07397}}
}

@article{deshmukh2025kascade,
  author        = {Deshmukh, Dhruv and Goyal, Saurabh and Kwatra, Nipun and Ramjee, Ramachandran},
  title         = {Kascade: A Practical Sparse Attention Method for Long-Context {LLM} Inference},
  journal       = {arXiv preprint \href{https://arxiv.org/abs/2512.16391}{arXiv:2512.16391}},
  year          = {2025}
}

@inproceedings{jones2026quoka,
  author        = {Dalton Jones and Junyoung Park and Matthew Morse and Mingu Lee and Chris Lott and M. Harper Langston},
  title         = {{QuoKA}: Query-Oriented {KV} Selection for Efficient {LLM} Prefill},
  booktitle     = {International Conference on Learning Representations},
  year          = {2026},
  url           = {https://openreview.net/forum?id=YS4N1YxXSM}
}

@inproceedings{nawrot2026sparsefrontier,
  author        = {Piotr Nawrot and Robert Li and Renjie Huang and Sebastian Ruder and Kelly Marchisio and Edoardo M. Ponti},
  title         = {The Sparse Frontier: Sparse Attention Trade-offs in Transformer {LLM}s},
  booktitle     = {Findings of the Association for Computational Linguistics: ACL 2026},
  year          = {2026},
  url           = {https://aclanthology.org/2026.findings-acl.1926/}
}

@inproceedings{hu2022lora,
  title     = {{LoRA}: Low-Rank Adaptation of Large Language Models},
  author    = {Hu, Edward J. and Shen, Yelong and Wallis, Phillip and Allen-Zhu, Zeyuan and Li, Yuanzhi and Wang, Shean and Wang, Lu and Chen, Weizhu},
  booktitle = {International Conference on Learning Representations},
  year      = {2022},
  url           = {https://openreview.net/forum?id=nZeVKeeFYf9}
}

@inproceedings{liu2024kivi,
  author = {Zirui Liu and Jiayi Yuan and Hongye Jin and Shaochen Zhong and Zhaozhuo Xu and Vladimir Braverman and Beidi Chen and Xia Hu},
  title = {{KIVI}: A Tuning-Free Asymmetric 2bit Quantization for {KV} Cache},
  booktitle = {Proceedings of the 41st International Conference on Machine Learning},
  year = {2024},
  url           = {https://proceedings.mlr.press/v235/liu24bz.html}
}

@inproceedings{li2024snapkv,
  author        = {Yuhong Li and Yingbing Huang and Bowen Yang and Bharat Venkitesh and Acyr Locatelli and Hanchen Ye and Tianle Cai and Patrick Lewis and Deming Chen},
  title         = {{SnapKV}: {LLM} Knows What You are Looking for Before Generation},
  booktitle     = {Advances in Neural Information Processing Systems},
  volume        = {37},
  year          = {2024},
  note          = {\href{https://arxiv.org/abs/2404.14469}{arXiv:2404.14469}}
}

@inproceedings{dong2024flexattention,
  author        = {Juechu Dong and Boyuan Feng and Driss Guessous and Yanbo Liang and Horace He},
  title         = {{FlexAttention}: A Programming Model for Generating Fused Attention Variants},
  booktitle     = {Proceedings of Machine Learning and Systems},
  volume        = {7},
  year          = {2025},
  note          = {\href{https://arxiv.org/abs/2412.05496}{arXiv:2412.05496}}
}

@misc{nvidia2026nemotron35,
  author        = {{NVIDIA}},
  title         = {{NVIDIA-Nemotron-3.5-Lightning-30B-A3B}},
  year          = {2026},
  howpublished  = {\url{https://huggingface.co/nvidia/NVIDIA-Nemotron-3.5-Lightning-30B-A3B-BF16}}
}

@article{ahmad2026spotattention,
  author        = {Huzama Ahmad and Se-Young Yun},
  title         = {{SpotAttention}: Plug-In Block-Sparse Routing for Pretrained Long-Context Transformers},
  journal       = {arXiv preprint \href{https://arxiv.org/abs/2606.22874}{arXiv:2606.22874}},
  year          = {2026}
}

@article{fan2026flashprefillv2,
  author        = {Qihang Fan and Huaibo Huang and Zhiying Wu and Bingning Wang and Ran He},
  title         = {{FlashPrefill V2}: Block-Sparse Prefill Attention for Long-Context {LLM} Serving},
  journal       = {arXiv preprint \href{https://arxiv.org/abs/2608.19758}{arXiv:2608.19758}},
  year          = {2026}
}

@article{fan2026uniprefill,
  author        = {Qihang Fan and Huaibo Huang and Zhiying Wu and Bingning Wang and Ran He},
  title         = {{UniPrefill}: Universal Long-Context Prefill Acceleration via Block-wise Dynamic Sparsification},
  journal       = {arXiv preprint \href{https://arxiv.org/abs/2605.06221}{arXiv:2605.06221}},
  year          = {2026}
}

@inproceedings{xiao2024efficient,
  title={Efficient Streaming Language Models with Attention Sinks},
  author={Xiao, Guangxuan and Tian, Yuandong and Chen, Beidi and Han, Song and Lewis, Mike},
  booktitle={The Twelfth International Conference on Learning Representations},
  year={2024},
  url           = {https://openreview.net/forum?id=NG7sS51zVF}
}

@article{su2024roformer,
  author        = {Jianlin Su and Murtadha Ahmed and Yu Lu and Shengfeng Pan and Wen Bo and Yunfeng Liu},
  title         = {{RoFormer}: Enhanced Transformer with Rotary Position Embedding},
  journal       = {Neurocomputing},
  volume        = {568},
  pages         = {127063},
  year          = {2024},
  url           = {https://doi.org/10.1016/j.neucom.2023.127063}
}

@inproceedings{xiao2024duoattention,
  author        = {Guangxuan Xiao and Jiaming Tang and Jingwei Zuo and Junxian Guo and Shang Yang and Haotian Tang and Yao Fu and Song Han},
  title         = {{DuoAttention}: Efficient Long-Context {LLM} Inference with Retrieval and Streaming Heads},
  booktitle     = {International Conference on Learning Representations},
  year          = {2025},
  note          = {\href{https://arxiv.org/abs/2410.10819}{arXiv:2410.10819}}
}

@inproceedings{wu2024retrievalhead,
  author        = {Wenhao Wu and Yizhong Wang and Guangxuan Xiao and Hao Peng and Yao Fu},
  title         = {Retrieval Head Mechanistically Explains Long-Context Factuality},
  booktitle     = {International Conference on Learning Representations},
  year          = {2025},
  note          = {\href{https://arxiv.org/abs/2404.15574}{arXiv:2404.15574}}
}
\bibliographystyle{iclr2027_conference}

\clearpage
\appendix
\raggedbottom
\section{Computational cost}
\label{app:notation}
\label{app:complexity}
\suppressfloats[t]

We consider a layer with $H_{kv}$ KV heads, each shared by a group $\mathcal H$ of $G=|\mathcal H|$ query heads, so it has $H_q=GH_{kv}$ query heads.
The prompt has $L$ tokens, queries and keys have dimension $d$, and each KV head has $C$ semantic blocks.
Dense attention costs $\Theta(H_qLd)$ per generated token and $\Theta(H_qL^2d)$ over a prefill.
Table~\ref{tab:stage-breakdown} reports the measured GPU time of each stage.

\paragraph{Building the index.}
The index is built once per layer and prompt.
With a fixed number of $k$-means iterations, fitting the centroids and assigning all prompt keys cost $O(H_{kv}LCd)$.
Normalizing the keys and packing the KV into block order each cost $O(H_{kv}Ld)$.
Appendix~\ref{app:implementation} gives the clustering settings.

\paragraph{Scoring and selection.}
Scoring $C$ centroids in each query head costs $O(H_qCd)$ per query position and $O(H_qLCd)$ over a prefill.
Selection runs once per KV head and uses a fixed number of threshold-refinement passes, costing $O(H_{kv}C)$ per position.
Per-token indexers such as DSA~\citep{deepseek2025v32} and MSA~\citep{lai2026msa} instead score all visible tokens, so their scoring remains quadratic over a prefill even with a small attention budget.

\paragraph{Attention.}
For each KV head, attention at position $t$ costs $O(Gd|\mathcal S_t|)$, where $\mathcal S_t$ is the set of positions read by that query (Section~\ref{sec:gqa-routing}).
Summing this cost over query positions and KV heads gives the full prefill cost.
The set includes selected blocks and positions read without selection, such as the current chunk during prefill or generated tokens during decoding.
Selecting whole blocks can exceed the budget $K$, and the initial dense queries read their full visible prefix.
Including all generated tokens makes decode cost grow with generation length.

\paragraph{Growth with prompt length.}
Our main configuration keeps the target mean block size $u$ fixed, so that $C\approx L/u$, and uses $K=L/16$.
Prefill therefore remains quadratic in prompt length.
More generally, prefill can become subquadratic if both $C$ and the average number of positions read per query grow sublinearly with $L$.

\paragraph{Memory.}
The index stores $O(H_{kv}Cd)$ centroid values and $O(H_{kv}L)$ block assignments.
Prompt KV is stored once, in block order.
Appendix~\ref{app:memory} reports peak memory, including temporary activations and kernel workspace.

\section{Implementation details}
\label{app:implementation}
\suppressfloats[t]

\subsection{Routing and clustering}
Clustering and selection use the pre-RoPE queries and keys of each head, 256-dimensional vectors normalized as whole vectors.
Each KV head has a separate prompt partition.
For a target mean block size $u$, the number of clusters is $C=\min\bigl(L,\ \max(128,\ 2^{\lceil\log_2(L/u)\rceil})\bigr)$.
Table~\ref{tab:clustering-settings} lists the resulting cluster counts at $u=2{,}048$.
At 128K, $L/u=64$, so $C=128$ and the mean block size is 1,024, half the target.
Centroids start from uniformly sampled keys and are refined for eight iterations on the sample in the table, with empty clusters re-seeded from it.
All prompt keys are assigned after fitting.

\begin{table}[t]
\caption{Clustering per KV head at $u=2{,}048$, at the four prompt lengths we report.}
\label{tab:clustering-settings}
\centering
\small
\begin{tabular}{lrrr}
\toprule
Prompt length & Clusters & Mean block size & Keys used to fit \\
\midrule
128K & 128 & 1{,}024 & 8{,}192 \\
256K & 128 & 2{,}048 & 8{,}192 \\
512K & 256 & 2{,}048 & 8{,}192 \\
1M & 512 & 2{,}046 & 16{,}384 \\
\bottomrule
\end{tabular}
\end{table}
In the fit-budget sweep, the sample count ranges from 2,048 to all prompt keys and the iteration count from two to sixteen.
MK-3 accuracy at 512K ranges from 90\% to 96\% over 100 items, while prefill latency varies by at most six percent.

For the PCA ablation, we fit a projection using about one million raw-corpus keys per layer and KV head, without subtracting their mean.
We apply the same projection to queries and keys, then normalize each vector.

\subsection{Selection and attention}
The prefill selector uses 16 threshold-refinement steps to approximate the score-ordered budget selection in Equation~\ref{eq:budget-selection}.
Its size penalty uses the visible block sizes $|\mathcal B'_{c,t}|$ of Equation~\ref{eq:gqa-score}, and blocks with no visible token are never selected.
The selected clusters are represented by a bitmask per query and KV head.
Counts and prefix sums convert this to cluster-major query lists and query-major merge slots, without materializing the full Boolean selection matrix or sorting partial outputs afterward.
Within each block of this cache, tokens stay in position order, and every block starts at a page boundary.
Each prefill chunk passes the exact count of preceding tokens in every cluster to FlashAttention-3, including a partially filled final page.
The selected history is therefore entirely before the chunk, and only the separate current-chunk call needs a triangular causal mask.
The GQA heads are folded into query rows for the history call.

Partial attention outputs are merged using their softmax normalizers.
If region $j$ returns output $o_j$ and log-sum-exp $\ell_j$, the combined output is $o=\sum_j \exp(\ell_j-\ell)\,o_j$ with $\ell=\operatorname{logsumexp}_j\ell_j$.
The merge is computed in FP32 and cast back to BF16, giving attention over the selected positions up to floating-point rounding.

The decode backend of the latency runs uses five launches per sparse layer, namely suffix write, query normalization, fused scoring and selection, split attention, and output merge.
Selection directly emits contiguous cluster ranges.
The attention kernel reads selected prompt ranges and the tiles of the generated suffix in the same launch, then merges their softmax statistics.
All live lengths are read on device and workspace is reused, avoiding per-step host reads and allocations.
The RULER and MRCR-8 runs and the 4B model's LongBench-v2 runs in Table~\ref{tab:quality-main} use a fused decode backend with more launches per layer, which selects the same clusters, under vLLM's piecewise CUDA graphs.

\begin{table}[t]
\centering
\small
\caption{Main Qwen3.5-4B runtime parameters. Benchmark-specific budgets are described in Section~\ref{sec:experiments}.}
\label{tab:implementation-params}
\begin{tabularx}{\linewidth}{lX}
\toprule
Quantity & Value \\
\midrule
Replaced layers & Eight full-attention layers of 32 \\
Query / KV heads & 16 / 4, head dimension 256 \\
Routing & Pre-RoPE Q/K, whole-vector normalization, identity map \\
Target mean block size & $u=2{,}048$, which sets the cluster count of Table~\ref{tab:clustering-settings} \\
Clustering & Eight iterations, $\min(L,\max(8{,}192,32C))$ sampled keys. The main RULER results average seeds 0--2, and other results use seed 0 unless stated otherwise (Appendix~\ref{app:seeds}) \\
Score parameters & $\tau=0.1$, $\beta=0.25$ \\
Length-controlled budget & $K=L/16$ for RULER and the main latency sweep \\
Query chunk & 8,192 positions \\
Dense prefix & A query chunk starting at position $4K$ (four times the budget) or earlier attends to its whole prefix without selection \\
Attended without selection ($\mathcal D_t$) & The visible part of the current chunk in prefill, every generated token in decoding \\
Numerics & BF16 attention, FP32 score aggregation and output merge \\
Short prompts & Sparse attention at every measured length, with the runtime's dense fallback for short prompts switched off \\
Decode graph & The whole decode step captured as one CUDA graph in the latency runs, for dense and ours \\
Prefill schedule & Sliced (32,768-position slices) at 1M, for the task evaluations above 256K, and for the main RULER-5 evaluations at every length (Appendix~\ref{app:prefill-schedules}). Single-pass for the latency sweep through 512K and the other task evaluations through 256K \\
\bottomrule
\end{tabularx}
\end{table}

\subsection{Compilation and decode graphs}
Measurements use \texttt{TORCHINDUCTOR\_DETERMINISTIC=1}, with combo kernels and their benchmarking disabled.
These settings reduce numerical differences between independent compilations.
Prefill uses piecewise graphs where applicable.
For the main decode latency results, both methods use vLLM's \texttt{FULL\_DECODE\_ONLY} mode, which captures the whole decode step as one CUDA graph.
Each sparse layer has a fixed-address arena containing prompt KV, centroids, posting lists, generated-token storage, and workspace.
The graph is captured at engine warm-up and reused after the prompt's index is copied into the arena.
One-time graph capture is excluded from steady decode latency for both methods.

\subsection{Prefill schedules}
\label{app:prefill-schedules}
An index fitted on the whole prompt needs all of a layer's prompt keys before that layer's queries can be routed.
The runtime therefore cannot use ordinary chunked prefill, which passes one token chunk through all layers before the next chunk.
Instead, every layer processes the whole prompt before the next layer starts, in one of two ways (Figure~\ref{fig:prefill-schedules}).
\emph{Single-pass} prefill runs the whole prompt through the model as one forward pass, so every projection and feed-forward call processes all prompt positions at once and holds full-length activations.
\emph{Sliced} prefill loops over the layers itself instead of calling the model's forward pass once, and processes each layer in slices of 32,768 positions.
A full-attention layer makes two passes over the slices.
The first projects the keys and values for all slices, then builds the index and packs the block-ordered KV cache.
The second projects the queries and the output gate, performs attention, and writes the outputs.
Linear-attention layers carry their recurrent and convolution states between slices, and feed-forward layers and normalization are sliced as well.
Both schedules compute the same attention up to numerical differences, with selection and attention in 8,192-position query chunks inside each layer, and differ in peak memory (Table~\ref{tab:memory-against-dense}).
Sliced prefill is required where single-pass prefill exceeds device memory, and Table~\ref{tab:implementation-params} lists which schedule each measurement uses.
The schedule is selected outside the compiled decode path, and elementwise operators are compiled separately for the sliced path.

\begin{figure}[htb]
\centering
\scriptsize
\setlength{\tabcolsep}{4pt}
\begin{tabular}{@{}p{0.30\linewidth}p{0.30\linewidth}p{0.34\linewidth}@{}}
\textbf{Dense attention, chunked} & \textbf{Ours, single-pass} & \textbf{Ours, sliced} \\[2pt]
\ttfamily for each 16K-token chunk:\newline
\hspace*{1em}for each layer:\newline
\hspace*{2em}Q, K, V <- project(chunk)\newline
\hspace*{2em}attend(chunk, KV so far)\newline
\hspace*{2em}feed-forward(chunk)
&
\ttfamily for each layer:\newline
\hspace*{1em}Q, K, V <- project(all)\newline
\hspace*{1em}fit index on all keys\newline
\hspace*{1em}for each 8K query chunk:\newline
\hspace*{2em}select, attend\newline
\hspace*{1em}feed-forward(all)
&
\ttfamily for each layer:\newline
\hspace*{1em}for each 32K slice:\newline
\hspace*{2em}K, V <- project(slice)\newline
\hspace*{1em}fit index on all keys\newline
\hspace*{1em}for each 32K slice:\newline
\hspace*{2em}Q <- project(slice)\newline
\hspace*{2em}for each 8K query chunk:\newline
\hspace*{3em}select, attend\newline
\hspace*{2em}feed-forward(slice)
\end{tabular}
\caption{The order of operations in one prefill, shown for a full-attention layer and its feed-forward block, where ``all'' is every prompt position. Linear-attention layers follow the same order without the index.}
\label{fig:prefill-schedules}
\end{figure}

\subsection{Memory}
\label{app:memory}
In the memory and latency runs, prompt KV is stored only in the block-ordered cache, with space reserved for generated tokens and workspace.
The 4B model's single-pass RULER runs at 128K and 256K also write prompt KV to vLLM's paged cache.
Table~\ref{tab:memory-against-dense} compares memory use with dense chunked prefill.
For our method, ``KV cache etc.'' is the allocated memory remaining after prefill, excluding weights.
``Transient'' is the additional allocation at the peak.

\begin{table}[t]
\centering
\small
\caption{Qwen3.5-4B memory on one H200, in GiB. Dense totals are derived request-sized estimates that exclude excess preallocated cache capacity, while sparse peaks are measured allocated memory, so dense and sparse rows are accounted differently.}
\label{tab:memory-against-dense}
\begin{tabular}{ll rrrr}
\toprule
Method & Length / schedule & Weights & KV cache etc. & Transient & Total \\
\midrule
Dense (derived) & 512K / chunked & 8.35 & 16.00 & 1.58 & 25.93 \\
Ours (measured) & 512K / single-pass & 8.35 & 20.84 & 53.50 & 82.69 \\
Ours (measured) & 512K / sliced & 8.35 & 20.84 & 11.91 & 41.09 \\
Ours, prefix index (measured) & 512K / chunked & 8.35 & 20.86 & 6.43 & 35.64 \\
Dense (derived) & 1M / chunked & 8.35 & 31.97 & 1.58 & 41.90 \\
Ours (measured) & 1M / sliced & 8.35 & 39.65 & 22.37 & 70.37 \\
Ours, prefix index (measured) & 1M / chunked & 8.35 & 39.72 & 11.89 & 59.96 \\
\bottomrule
\end{tabular}
\end{table}

The additional memory over dense chunked prefill has two sources.
The sliced schedule of Appendix~\ref{app:prefill-schedules} holds the residual stream for the whole prompt and, until a layer's index is built, that layer's keys and values, which grow with prompt length.
The KV-outer attention of Section~\ref{sec:implementation} holds, for one query chunk and KV head at a time, the gathered queries and their partial outputs, which grow with the query chunk and the number of blocks its queries selected.
The 512K latency runs use single-pass prefill, whose peak is listed separately and is set by the full-length feed-forward intermediates.
Both methods fit on one H200 at every length measured.
Sparsity does not reduce memory here, because the full prompt KV is kept.

The prefix-index rows fit centroids on the prompt's first 64K tokens and assign later keys to them as they arrive.
This allows ordinary chunked prefill with 8K-token chunks and no full-length activations.
The remaining memory overhead comes from the output buffer and the attention kernels' per-chunk buffers.
This schedule is about a sixth slower than sliced prefill.
The first-64K partition gives similar accuracy to the full-prompt partition on the four RULER tasks in Table~\ref{tab:centroid-source}.
Running chunked prefill with this index gives 93.0 on MK-3 at 512K, compared with 95.0 for the first-64K row in Table~\ref{tab:centroid-source}.
The reported results use the full-prompt index.

\section{Measurement protocol}
\label{app:protocol}
\suppressfloats[t]

\subsection{Latency}
All headline latency measurements use one request on one H200 and BF16 weights.
Both methods use FlashAttention-3 for prefill, and our decoding uses the fused backend of Appendix~\ref{app:implementation}.
Each latency measurement uses one warm-up followed by three measured repetitions.
Each request generates 64 tokens for prompts through 512K and 32 tokens at 1M.
For each repetition, decode latency is the time between the first and last generated tokens divided by the number of tokens after the first.
We report the median over the three repetitions.
For the 4B model, our method is timed on two different prompts of each length through 512K and on one prompt at 1M.
Dense attention is timed on two prompts at 512K and on one prompt at 128K, 256K, and 1M.
Where there are two prompts, we report their mean.
The 35B-A3B model is timed on one prompt at every length under the same protocol.
Both decode paths use full-step CUDA graphs.
We measure our method's latency using single-pass prefill through 512K on the 4B model and through 256K on the 35B-A3B model, with sliced prefill at longer lengths.
The dense baseline uses vLLM's chunked prefill.
Layer timings sum kernels attributed to the attention backend, with decode accounting over the final 32 generated steps.
They are profiler totals, not whole-model wall-clock measurements.

\subsection{Quality}
RULER uses 100 cached prompts per task and length, shared by the two Qwen models, with greedy decoding and thinking disabled.
On the 4B model, our main RULER-5 evaluations and all task evaluations above 256K use sliced prefill (Table~\ref{tab:implementation-params}).
All sparse task evaluations of the 35B-A3B model use sliced prefill.

MRCR-8 uses revision \texttt{f4c69fa} (8 December 2025) and the model's thinking sampling preset.
We keep the official length-bin boundaries and assign prompts using their length in Qwen3.5 tokens after the chat template.
The bins contain 100 items at 128K and 512K and 99 at 256K.
The generation limit accommodates the longest answer and a 32,768-token thinking budget.

LongBench-v2 retains 470 items whose raw prompts fit within 983,040 tokens.
Thinking is enabled, with at most 40,960 generated tokens.
Sampling uses temperature 1.0, top-$p$ 0.95, top-$k$ 20, min-$p$ 0, and presence penalty 1.5.

The main MRCR-8 and LongBench-v2 results average three runs, each with its own sampling seed and, for our method, its own clustering seed.
Budgets are defined in Section~\ref{sec:experiments}.

\subsection{Prompt lengths}
\label{app:prompt-lengths}
The prompts come from lm\_eval's port of the RULER generator, with NVIDIA's filling for common-word extraction beyond 128K, where the port's word list runs out.
The measured median RULER-5 prompt lengths are within five percent of nominal at 128K and two percent from 256K onward, except multi-value retrieval at 1M, whose generator runs out of material at about 647K tokens.
Of the other tasks at 512K, frequent-word extraction is the one the generator sizes by words rather than tokens, and its prompts run from 418K to 505K tokens with a median of 476K. The budget follows the nominal length.
SQuAD QA is not in RULER-5 because filling it repeats the answer paragraph, about two and a half times at 512K and five times at 1M, so raising the nominal length also changes how often the answer appears.

\subsection{Clustering seeds and seeding}
\label{app:seeds}
For both Qwen models, the main RULER-5 results average three clustering seeds with random initialization (Tables~\ref{tab:main-seeds} and~\ref{tab:ruler-by-task}).
Appendix comparisons use seed 0 unless stated otherwise.
When the main configuration is rerun alongside variants, we report its score from that run, which can differ across tables because the fitted partition changes.
When it is included only as a reference, we use its three-seed mean.

Random initialization draws from a sample of 8,192 keys, and $k$-means++ uses $D^2$ sampling on the same sample.
Greedy farthest-point seeding uses 65,536 keys for Qwen seed 0 and 262,144 for seeds 1 and 2, and 1,024 keys per centroid on Nemotron.
Neither alternative gives a consistent improvement (Table~\ref{tab:seeding}), so we retain random initialization.

Across three seeds on the same 512K items, MK-3 accuracy varies by one point and MK-2 by two, while up to 8\% of the items in either task change between correct and incorrect.

\begin{table}[t]
\caption{The clustering seed and the seeding procedure, with RULER multi-key accuracy in percent at $K=L/16$ and $u=2{,}048$, 100 items per task and length.}
\label{tab:seeding}
\centering
\small
\begin{tabular}{l l l l c}
\toprule
Model & Task & Length & Seeding & Mean (s.d.) \\
\midrule
Qwen3.5-4B & MK-2 & 512K & random & 95.0 (1.0) \\
 & & & farthest & 92.7 (1.5) \\
 & & & $k$-means++ & 91.7 (1.5) \\
\addlinespace
 & MK-3 & 512K & random & 92.7 (0.6) \\
 & & & farthest & 94.3 (1.5) \\
 & & & $k$-means++ & 92.0 (3.6) \\
\addlinespace
Nemotron 3.5 & MK-3 & 256K & random & 73 (1 seed) \\
 & & & farthest & 86 (1 seed) \\
\addlinespace
 & & 512K & random & 84 (1 seed) \\
 & & & farthest & 80 (1 seed) \\
\bottomrule
\end{tabular}
\end{table}

\begin{table}[t]
\caption{RULER results of our method in the main table, Qwen3.5-4B at $K=L/16$ and $u=2{,}048$, 100 items per task and length. Each entry is the mean (sample s.d.) over three clustering seeds. RULER-5 is the mean over the five tasks, and its s.d. is computed over the three seeds' five-task means.}
\label{tab:main-seeds}
\centering
\small
\begin{tabular}{l ccccc c}
\toprule
Prompt & MK-3 & MK-2 & MV & CWE & HotpotQA & RULER-5 \\
\midrule
128K & 100.0 (0.0) & 98.7 (1.5) & 99.8 (0.3) & 78.3 (0.5) & 57.3 (1.2) & 86.8 (0.3) \\
256K & 97.7 (0.6) & 91.0 (2.0) & 99.3 (0.4) & 70.3 (1.4) & 55.0 (1.0) & 82.7 (0.1) \\
512K & 92.7 (0.6) & 95.0 (1.0) & 100.0 (0.0) & 64.3 (0.9) & 57.0 (1.0) & 81.8 (0.5) \\
1M & 28.3 (1.5) & 68.0 (2.0) & 100.0 (0.0) & 29.4 (1.2) & 50.0 (1.0) & 55.2 (0.8) \\
\bottomrule
\end{tabular}

\end{table}

\Needspace{0.55\textheight}
\section{Quality in detail}
\label{app:quality-4b}
\suppressfloats[t]

\subsection{RULER results by task}
\label{app:ruler-by-task}
\begin{table}[htb]
\caption{Every RULER task on Qwen3.5-4B at a nominal 512K, in percent, 100 items per task, $K=L/16$ except the last row, which halves it. Ours on the starred tasks and SQuAD is the mean over three clustering seeds, on the others one seed. The second row is the median length of each task's prompts in thousands of tokens (Appendix~\ref{app:prompt-lengths}). S1--S3 are the single-needle tasks, MK-1 to MK-3 multi-key, MQ multi-query, MV multi-value, VT variable tracking, CWE and FWE common- and frequent-word extraction, SQuAD the QA task with its paragraphs repeated, and Hotpot HotpotQA. The starred tasks form RULER-5.}
\label{tab:ruler-all-tasks}
\centering
\small
\setlength{\tabcolsep}{3.4pt}
\resizebox{\linewidth}{!}{%
\begin{tabular}{l rrr rrr r r r rr r r r}
\toprule
& S1 & S2 & S3 & MK-1 & MK-2$^*$ & MK-3$^*$ & MQ & MV$^*$ & VT & CWE$^*$ & FWE & SQuAD & Hotpot$^*$ & Mean \\
Prompt (K tokens) & 524 & 524 & 524 & 524 & 524 & 523 & 524 & 524 & 524 & 523 & 476 & 523 & 517 & \\
\midrule
Dense & 100.0 & 100.0 & 100.0 & 100.0 & 99.0 & 96.0 & 99.8 & 100.0 & 100.0 & 63.0 & 99.3 & 76.1 & 56.0 & 91.5 \\
Ours & 100.0 & 100.0 & 99.0 & 100.0 & 95.0 & 92.7 & 100.0 & 100.0 & 100.0 & 64.3 & 99.0 & 79.1 & 57.0 & 91.2 \\
Ours, $K=L/32$ & 100.0 & 100.0 & 99.0 & 99.0 & 76.7 & 82.7 & 99.5 & 99.5 & 100.0 & 63.2 & 93.7 & 74.3 & 52.0 & 87.7 \\
\bottomrule
\end{tabular}}
\end{table}

On Qwen3.5-4B at 512K, our mean accuracy over all thirteen RULER tasks remains close to dense attention (Table~\ref{tab:ruler-all-tasks}).
The largest losses occur on multi-key retrieval, where MK-2 identifies needles by words and MK-3 by UUIDs.
Halving the budget to $K=L/32$ further reduces multi-key accuracy, with smaller drops on frequent-word extraction and the two QA tasks.
Most tasks outside RULER-5 are near ceiling.
RULER-5 focuses on the harder tasks and includes multi-value retrieval to check recovery of multiple answers.

\begin{table}[t]
\caption{RULER-5 by task and length on the three models, in percent, on 100 examples per task at $K=L/16$. For the Qwen models, the ours rows are means over three clustering seeds, and the 35B-A3B model also has multi-key rows at half and a quarter of the budget at 512K. Nemotron is measured at 512K with one seed, and its last row doubles the budget. Common-word extraction beyond 128K is on the filled prompts, and the 1M multi-value prompts are about 647K tokens long (Appendix~\ref{app:prompt-lengths}).}
\label{tab:ruler-by-task}
\centering
\scriptsize
\setlength{\tabcolsep}{3.2pt}
\begin{tabular}{r r l rrrrr r}
\toprule
Prompt & $K$ & Method & Multi-key-3 & Multi-key-2 & Multi-value & Common-word & HotpotQA & Mean \\
\midrule
\multicolumn{9}{l}{\textit{Qwen3.5-4B}} \\
\midrule
\multirow{3}{*}{128K} & \multirow{3}{*}{8,192}
& Dense & 100.0 & 100.0 & 100.0 & 78.7 & 58.0 & 87.3 \\
&& Ours & 100.0 & 98.7 & 99.8 & 78.3 & 57.3 & 86.8 \\
&& \textit{2B dense} & 100.0 & 100.0 & 99.8 & 77.8 & 52.0 & 85.9 \\
\midrule
\multirow{3}{*}{256K} & \multirow{3}{*}{16,384}
& Dense & 100.0 & 100.0 & 100.0 & 68.3 & 55.0 & 84.7 \\
&& Ours & 97.7 & 91.0 & 99.3 & 70.3 & 55.0 & 82.7 \\
&& \textit{2B dense} & 97.0 & 99.0 & 100.0 & 46.7 & 46.0 & 77.7 \\
\midrule
\multirow{3}{*}{512K} & \multirow{3}{*}{32,768}
& Dense & 96.0 & 99.0 & 100.0 & 63.0 & 56.0 & 82.8 \\
&& Ours & 92.7 & 95.0 & 100.0 & 64.3 & 57.0 & 81.8 \\
&& \textit{2B dense} & 70.0 & 98.0 & 98.2 & 19.1 & 47.0 & 66.5 \\
\midrule
\multirow{3}{*}{1M} & \multirow{3}{*}{65,536}
& Dense & 40.0 & 89.0 & 100.0 & 35.1 & 52.0 & 63.2 \\
&& Ours & 28.3 & 68.0 & 100.0 & 29.4 & 50.0 & 55.2 \\
&& \textit{2B dense} & 20.0 & 89.0 & 98.0 & 8.3 & 29.0 & 48.9 \\
\midrule
\multicolumn{9}{l}{\textit{Qwen3.5-35B-A3B}} \\
\midrule
\multirow{2}{*}{128K} & \multirow{2}{*}{8,192}
& Dense & 100.0 & 100.0 & 100.0 & 79.7 & 63.0 & 88.5 \\
&& Ours & 98.7 & 100.0 & 99.0 & 82.2 & 63.7 & 88.7 \\
\midrule
\multirow{2}{*}{256K} & \multirow{2}{*}{16,384}
& Dense & 100.0 & 99.0 & 100.0 & 69.5 & 60.0 & 85.7 \\
&& Ours & 96.3 & 99.0 & 99.4 & 68.6 & 59.3 & 84.5 \\
\midrule
\multirow{2}{*}{512K} & \multirow{2}{*}{32,768}
& Dense & 100.0 & 97.0 & 100.0 & 68.0 & 57.0 & 84.4 \\
&& Ours & 98.7 & 97.7 & 98.8 & 67.2 & 58.7 & 84.2 \\
& 16,384 & Ours, $K=L/32$ & 80.0 & 96.0 & -- & -- & -- & -- \\
& 8,192 & Ours, $K=L/64$ & 25.0 & 68.0 & -- & -- & -- & -- \\
\midrule
\multirow{2}{*}{1M} & \multirow{2}{*}{65,536}
& Dense & 85.0 & 85.0 & 100.0 & 48.2 & 48.0 & 73.2 \\
&& Ours & 79.3 & 88.7 & 100.0 & 44.9 & 56.7 & 73.9 \\
\midrule
\multicolumn{9}{l}{\textit{Nemotron 3.5 Lightning}} \\
\midrule
\multirow{3}{*}{512K} & 32,768 & Dense & 100.0 & 99.0 & 100.0 & 95.7 & 61.0 & 91.1 \\
&  & Ours & 84.0 & 49.0 & 71.5 & 92.4 & 53.0 & 70.0 \\
& 65,536 & Ours, $K=L/8$ & 90.0 & 81.0 & 95.5 & 95.1 & 56.0 & 83.5 \\
\bottomrule
\end{tabular}

\end{table}

The italic rows under the 4B model are the dense Qwen3.5-2B model on the same items, a reference for how much a smaller dense model loses.
On Qwen3.5-4B, sparsification reduces accuracy most on the two multi-key tasks from 256K onward, especially MK-2 (Table~\ref{tab:ruler-by-task}).
HotpotQA and common-word extraction remain close to dense attention through 512K.
Both fall a few points below dense attention at 1M.

\subsection{Positional scaling beyond the native context}
\label{app:positional-scaling}

YaRN changes position interpolation and attention scaling~\citep{peng2024yarn}.
We set YaRN's attention factor to $\alpha=1/\mathrm{mscale}(s)$, with $\mathrm{mscale}(s)=0.1\ln s+1$ ($\alpha=0.878$ at $s=4$). The runtime multiplies the rotary cosine and sine tables by $\alpha\,\mathrm{mscale}(s)=1$, which cancels the attention scaling while retaining interpolation.
Table~\ref{tab:positional-scaling} compares these settings on matched 512K prompts on the 4B model. The 35B-A3B model uses the chosen setting without a sweep of its own.
Unless stated otherwise, the interpolation factor is four.

\begin{table}[t]
\caption{Qwen3.5-4B and Qwen3.5-2B under alternative context-extension settings at 512K, with multi-key and multi-value accuracy in percent, $n=100$ per task, one clustering seed. Unqualified YaRN rows use factor four, and interpolation-only cancels attention-temperature scaling.}
\label{tab:positional-scaling}
\centering
\small
\setlength{\tabcolsep}{4pt}
\begin{tabular}{l ccc}
\toprule
Configuration & MK-3 & MK-2 & MV \\
\midrule
Dense, no extension        & 56.0 & 91.0 & 98.8 \\
Dense, YaRN                & 89.0 & 94.0 & 99.5 \\
Dense, YaRN $s=2$          & 74.0 & 95.0 & 100.0 \\
Dense, $s=2$, interp only  & 77.0 & 98.0 & 100.0 \\
\textbf{Dense, interpolation only} & \textbf{96.0} & \textbf{99.0} & \textbf{100.0} \\
\midrule
Ours, no extension & 46.0 & 66.0 & 99.8 \\
Ours, YaRN                 & 83.0 & 81.0 & 99.0 \\
\textbf{Ours, interpolation only} & \textbf{94.0} & \textbf{93.0} & \textbf{100.0} \\
\midrule
Qwen3.5-2B dense, YaRN & 44.0 & 87.0 & 95.8 \\
\textbf{Qwen3.5-2B dense, interpolation only} & \textbf{70.0} & \textbf{98.0} & \textbf{98.3} \\
\bottomrule
\end{tabular}

\end{table}

We chose the setting using dense attention's five-task mean, with QA and common-word extraction evaluated on the port's capped prompts (Appendix~\ref{app:prompt-lengths}).
It also gives the best multi-key scores in Table~\ref{tab:positional-scaling}.
The bold dense row is the main table's dense measurement, while the bold sparse row is an earlier single-seed run of the main configuration.
Removing the extra attention scaling improves both dense and sparse attention on the 4B model.
The dense Qwen3.5-2B reference also performs best with this setting and uses it throughout.

\subsection{Our method with a fixed initial block}
\label{app:sink}
We test a fixed block of the first 1,024 prompt positions to retain attention sinks, as in other sparse attention methods~\citep{xiao2024efficient,jiang2024minference}.
These positions are excluded from clustering, and every query includes the block in both prefill and decoding.
The block counts toward $K$, leaving the remaining budget for clusters.
All other settings follow the main configuration, which omits this block.
Section~\ref{sec:results-positional} applies the same block to the positional baseline (Table~\ref{tab:positional-points}).

Tables~\ref{tab:sink-ruler} and~\ref{tab:sink-mrcr-longbench} compare our method with and without the initial block on the same items.
On Qwen3.5-4B, the block recovers most of the 512K MK-3 loss, changes MK-2 little, and slightly improves RULER-5 across lengths.
RULER changes are small on Qwen3.5-35B-A3B.
On Nemotron 3.5 Lightning, it improves multi-value retrieval but lowers MK-2 accuracy.
Aggregate LongBench-v2 accuracy rises slightly on both Qwen models.
At 512K, MRCR-8 improves on the 4B model and declines on the 35B-A3B model.

\begin{table}[t]
\caption{RULER-5 by length and task at 512K, without and with the initial block, on Qwen3.5-4B, Qwen3.5-35B-A3B, and Nemotron 3.5 Lightning. Qwen rows average three clustering seeds.}
\label{tab:sink-ruler}
\centering
\small
\setlength{\tabcolsep}{3.2pt}
\begin{tabular}{l cccc ccccc}
\toprule
 & \multicolumn{4}{c}{RULER-5} & \multicolumn{5}{c}{Tasks at 512K} \\
\cmidrule(lr){2-5} \cmidrule(lr){6-10}
Method & 128K & 256K & 512K & 1M & MK-3 & MK-2 & MV & CWE & HotpotQA \\
\midrule
\multicolumn{10}{l}{\textit{Qwen3.5-4B}} \\
Dense & 87.3 & 84.7 & 82.8 & 63.2 & 96.0 & 99.0 & 100.0 & 63.0 & 56.0 \\
Ours & 86.8 & 82.7 & 81.8 & 55.2 & 92.7 & 95.0 & 100.0 & 64.3 & 57.0 \\
Ours with initial block & 87.0 & 83.6 & 82.5 & 56.6 & 95.0 & 94.7 & 99.2 & 66.2 & 57.3 \\
\midrule
\multicolumn{10}{l}{\textit{Qwen3.5-35B-A3B}} \\
Dense & 88.5 & 85.7 & 84.4 & 73.2 & 100.0 & 97.0 & 100.0 & 68.0 & 57.0 \\
Ours & 88.7 & 84.5 & 84.2 & 73.9 & 98.7 & 97.7 & 98.8 & 67.2 & 58.7 \\
Ours with initial block & 88.6 & 84.5 & 84.0 & 74.5 & 99.3 & 98.0 & 98.1 & 66.5 & 58.3 \\
\midrule
\multicolumn{10}{l}{\textit{Nemotron 3.5 Lightning}} \\
Dense & -- & -- & 91.1 & -- & 100.0 & 99.0 & 100.0 & 95.7 & 61.0 \\
Ours & -- & -- & 70.0 & -- & 84.0 & 49.0 & 71.5 & 92.4 & 53.0 \\
Ours with initial block & -- & -- & 74.7 & -- & 81.0 & 42.0 & 99.0 & 96.5 & 55.0 \\
Ours, $K=L/8$ & -- & -- & 83.5 & -- & 90.0 & 81.0 & 95.5 & 95.1 & 56.0 \\
Ours, $K=L/8$, with initial block & -- & -- & 83.3 & -- & 91.0 & 73.0 & 99.8 & 96.8 & 56.0 \\
\bottomrule
\end{tabular}

\end{table}

\begin{table}[htb]
\caption{MRCR-8 and LongBench-v2 of Qwen3.5-4B and Qwen3.5-35B-A3B without and with the initial block, the mean over three runs with the standard deviation across runs in parentheses.}
\label{tab:sink-mrcr-longbench}
\centering
\small
\resizebox{\linewidth}{!}{\begin{tabular}{l ccc ccc}
\toprule
 & \multicolumn{3}{c}{MRCR-8} & \multicolumn{3}{c}{LongBench-v2} \\
\cmidrule(lr){2-4} \cmidrule(lr){5-7}
Method & 128K & 256K & 512K & $\le$256K & $>$256K & all \\
\midrule
\multicolumn{7}{l}{\textit{Qwen3.5-4B}} \\
Dense & 0.797 (0.053) & 0.574 (0.010) & 0.412 (0.027) & 54.3 (1.1) & 64.2 (2.6) & 55.7 (0.7) \\
Ours & 0.788 (0.056) & 0.561 (0.029) & 0.399 (0.013) & 53.9 (1.0) & 57.7 (2.3) & 54.5 (0.6) \\
Ours with initial block & 0.783 (0.024) & 0.561 (0.068) & 0.415 (0.022) & 55.5 (0.9) & 57.2 (3.1) & 55.7 (0.4) \\
\midrule
\multicolumn{7}{l}{\textit{Qwen3.5-35B-A3B}} \\
Dense & 0.846 (0.012) & 0.655 (0.009) & 0.564 (0.056) & 64.4 (0.8) & 66.2 (2.3) & 64.7 (0.6) \\
Ours & 0.844 (0.016) & 0.668 (0.013) & 0.504 (0.063) & 63.1 (0.9) & 64.7 (0.9) & 63.3 (0.9) \\
Ours with initial block & 0.885 (0.019) & 0.655 (0.020) & 0.489 (0.021) & 64.7 (1.2) & 65.2 (4.3) & 64.8 (1.5) \\
\bottomrule
\end{tabular}
}
\end{table}

All rows use $K=L/16$ except the Nemotron rows at $K=L/8$.
Common-word extraction beyond 128K is on the filled prompts (Appendix~\ref{app:prompt-lengths}).

\Needspace{0.65\textheight}
\section{Latency in detail}
\label{app:latency-4b}
\suppressfloats[t]

\subsection{Latency by length}
\label{app:runtime-breakdown}
\begin{figure}[t]
\centering
\includegraphics[width=0.9\linewidth]{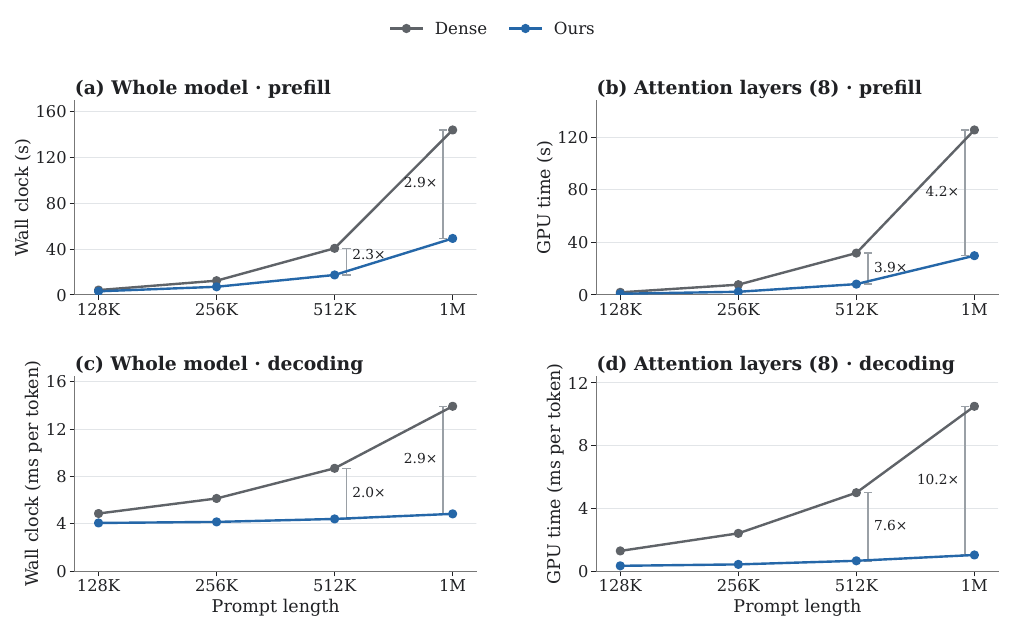}
\caption{Qwen3.5-4B latency on one H200 at $K=L/16$. The columns show whole-model wall-clock latency (left) and GPU time over the eight full-attention layers (right). The rows show prefill (top) and decoding (bottom).}
\label{fig:latency-detailed}
\end{figure}

\begin{table}[t]
\caption{Qwen3.5-4B latency on one H200. Whole-model columns are wall-clock measurements, the attention-layer columns profiler totals over the eight full-attention layers. The last rows are the dense reference model, Qwen3.5-2B.}
\label{tab:runtime}
\centering
\small
\setlength{\tabcolsep}{4.0pt}
\resizebox{\linewidth}{!}{\begin{tabular}{r cc cc cc cc}
\toprule
& \multicolumn{4}{c}{Prefill (s)} & \multicolumn{4}{c}{Decode (ms/token)} \\
\cmidrule(lr){2-5} \cmidrule(lr){6-9}
& \multicolumn{2}{c}{Whole model} & \multicolumn{2}{c}{Attention layers (8)} & \multicolumn{2}{c}{Whole model} & \multicolumn{2}{c}{Attention layers (8)} \\
Prompt & Dense & Ours & Dense & Ours & Dense & Ours & Dense & Ours \\
\midrule
128K & 4.18 & 3.11 & 1.92 & 0.84 & 4.87 & 4.07 & 1.29 & 0.34 \\
256K & 12.37 & 7.06 & 7.75 & 2.34 & 6.14 & 4.16 & 2.41 & 0.43 \\
512K & 40.69 & 17.39 & 31.81 & 8.18 & 8.69 & 4.41 & 5.00 & 0.66 \\
1M & 144.08 & 49.30 & 125.47 & 29.79 & 13.93 & 4.84 & 10.51 & 1.03 \\
\midrule
\multicolumn{9}{l}{\textit{Reference, Qwen3.5-2B with dense attention, whole model}} \\
128K & 1.72 & -- & -- & -- & 2.89 & -- & -- & -- \\
256K & 4.99 & -- & -- & -- & 3.48 & -- & -- & -- \\
512K & 16.01 & -- & -- & -- & 4.73 & -- & -- & -- \\
1M & 56.16 & -- & -- & -- & 7.48 & -- & -- & -- \\
\bottomrule
\end{tabular}
}
\end{table}

Figure~\ref{fig:latency-detailed} and Table~\ref{tab:runtime} report prefill and decode latency for Qwen3.5-4B.
Our method uses single-pass prefill through 512K and sliced prefill at 1M.
The nominal 1M latency prompt contains 1,047,552 tokens.
Appendix~\ref{app:protocol} gives the measurement protocol.

Appendix~\ref{app:architectures} reports the corresponding results for Qwen3.5-35B-A3B (Figure~\ref{fig:qwen35b-latency}) and Nemotron 3.5 Lightning (Figure~\ref{fig:nemotron-latency}).

\subsection{Prefill kernel breakdown}
\label{app:pipeline-stages}

\begin{table}[t]
\caption{GPU time in seconds for a 512K prefill on Qwen3.5-4B.
The attention rows contain only FlashAttention-3 kernels, with scoring, selection, gathering, and merging reported separately.
Totals come from profiling rather than the wall-clock medians in Table~\ref{tab:runtime}.}
\label{tab:stage-breakdown}
\centering
\small
\begin{tabular}{lrr}
\toprule
 & Dense & Ours \\
\midrule
Attention (8 attention layers) & 31.4 & 6.4 \\
\quad selected blocks & & 4.02 \\
\quad dense prefix & & 2.02 \\
\quad current chunk & & 0.34 \\
Scoring, selection, index build (ours) & 0.0 & 0.3 \\
\quad score the centroids and select & & 0.24 \\
\quad $k$-means fit & & 0.02 \\
\quad index build & & 0.02 \\
Gather, merge and other overhead (ours) & 0.0 & 1.5 \\
\quad gather and index operations & & 0.72 \\
\quad merge of partial outputs & & 0.38 \\
\quad masks, casts and copies & & 0.38 \\
Linear projections (all 32 layers) & 6.4 & 6.4 \\
Linear-attention layers (24 GDN) & 1.6 & 1.8 \\
Norms, activations, RoPE & 0.7 & 0.7 \\
Other & 0.1 & 0.0 \\
\midrule
Total & 40.3 & 17.1 \\
\bottomrule
\end{tabular}

\end{table}

GPU kernels are grouped by the operator that launches them.
At 512K, the first 17 of 64 query chunks use dense attention.
The other chunks attend to selected blocks and the visible part of their own chunk.

\subsection{Dense-prefix cutoff}
\label{app:dense-prefix}

The main configuration uses dense attention for a query chunk whose zero-based starting position satisfies $p\le4K$, four times the token budget.
We vary this cutoff to $p\le xK$ for $x\in\{4,2,1\}$ while keeping the token budget and other settings fixed.
Because the boundary is inclusive and selection is decided per chunk, $xK$ is the cutoff for chunk starts rather than the exact number of densely processed queries.

\begin{table}[t]
\caption{Effect of the dense-prefix cutoff on Qwen3.5-4B at 512K and $K=32{,}768$.
Timings are GPU seconds for one prefill, and accuracy is measured on 100 examples per task.}
\label{tab:dense-prefix}
\centering
\small
\setlength{\tabcolsep}{3pt}
\begin{tabular}{lrrrrrrrrr}
\toprule
& \multicolumn{7}{c}{GPU seconds of one 512K prefill} & \multicolumn{2}{c}{Accuracy (\%)} \\
\cmidrule(lr){2-8} \cmidrule(lr){9-10}
Setting & \shortstack[r]{Selected\\blocks} & \shortstack[r]{Dense\\prefix} & \shortstack[r]{Own\\chunk} & Attention & Scoring & \shortstack[r]{Gather,\\merge} & Total & MK-3 & MK-2 \\
\midrule
Dense attention & --- & --- & --- & 31.44 & --- & --- & 40.30 & 96.0 & 99.0 \\
$x=4$ (ours) & 4.02 & 2.03 & 0.34 & 6.38 & 0.24 & 1.48 & 17.07 & 96.0 & 97.0 \\
$x=2$ & 5.02 & 0.53 & 0.40 & 5.95 & 0.27 & 1.81 & 17.03 & 94.0 & 94.0 \\
$x=1$ & 5.65 & 0.16 & 0.43 & 6.24 & 0.29 & 2.10 & 17.64 & 93.0 & 93.0 \\
\bottomrule
\end{tabular}

\end{table}

Reducing the cutoff substantially lowers the dense-prefix attention time (Table~\ref{tab:dense-prefix}).
However, the affected queries now require block selection and grouped attention, increasing selected-block computation, query gathering, and output merging.
At $x=2$, these added costs largely offset the savings, leaving total prefill time nearly unchanged.
At $x=1$, total time increases.
Neither smaller cutoff consistently improves the two retrieval scores, so we retain $x=4$.

The dense reference is from Table~\ref{tab:stage-breakdown}.
The quality comparisons use one clustering seed on the same examples, whereas the main RULER-5 results average three seeds.

\subsection{Ablation of the kernel design}
\label{app:ablation}

Table~\ref{tab:ablation} compares three implementation choices from Section~\ref{sec:implementation} with simpler alternatives.
All runs use one GPU and the 512K sliced prefill of Qwen3.5-4B at $K=32{,}768$, with block selection and the rest of the runtime unchanged.
In Q-outer execution, each tile of 32 queries reads the union of their selected blocks, with each query masked to its own selection.
This takes more than twice as long as KV-outer execution and is slower than dense attention.
Blocks are read again for every tile that selects them, and masked query--block pairs still incur computation.
Gathering selected blocks from prompt-order KV for every query chunk adds about 9\% to prefill time.
The cluster-ordered cache avoids this repeated gathering.
Replacing the fused selection kernel with a matrix product and a sort costs well under a second.

\begin{table}[t]
\caption{Ablation of the kernel design on Qwen3.5-4B at 512K and $K=32{,}768$, with the prefill and decode times of each execution.}
\label{tab:ablation}
\centering
\small
\begin{tabular}{l rr}
\toprule
Execution & Prefill (s) & Decode (ms/token) \\
\midrule
Dense attention & 40.69 & 9.27 \\
Ours: KV-outer over one cluster-ordered copy & 18.51 & 4.49 \\
Selected blocks gathered from prompt order every chunk & 20.24 & 4.45 \\
Same selection, Q-outer instead of KV-outer & 47.56 & 4.29 \\
Selection by a GEMM and a sort & 19.09 & 4.56 \\
\bottomrule
\end{tabular}

\end{table}

We also time the attention kernel on one chunk of the same prefill.
With the same per-query selection, KV-outer execution using FlashAttention-3 takes about 21\,ms, compared with 245\,ms for FlexAttention~\citep{dong2024flexattention} using a block mask.
Most 128-by-128 blocks are partially selected, so FlexAttention processes nearly all of them.

\subsection{Indexing older generated tokens}
\label{app:suffix-flush}
\begin{figure}[t]
\centering
\includegraphics[width=0.6\linewidth]{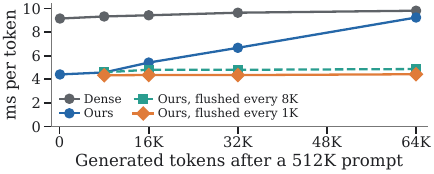}
\caption{Decode latency of Qwen3.5-4B after a 512K prompt, index updates included.}
\label{fig:generated-length}
\end{figure}

An optional flushing policy initially reads generated tokens exactly.
When the unindexed suffix reaches $W+F$ tokens, the oldest $F$ are assigned to the existing centroids and added to the index without refitting.
The newest $W$ remain outside the index.
The cache buffers stay at fixed addresses, so the captured decode graph remains valid.
Each point in Figure~\ref{fig:generated-length} averages the decode steps since the previous point, and all times include index updates.

\begin{table}[t]
\caption{Qwen3.5-4B mean decode time over the full generation after a nominal 1M prompt, in milliseconds per token.}
\label{tab:generated-length-1m}
\centering
\small
\begin{tabular}{l rr}
\toprule
 & 64 generated & 65,536 generated \\
\midrule
Dense & 15.39 & 16.06 \\
Ours & 4.81 & 7.85 \\
Ours, flushed every 8K ($W=2{,}048$) & 4.83 & 5.09 \\
Ours, flushed every 1K ($W=512$) & 4.82 & 4.82 \\
\bottomrule
\end{tabular}
\end{table}

\begin{table}[t]
\caption{Qwen3.5-4B LongBench-v2 accuracy (\%) without and with flushing ($F=1{,}024$, $W=512$), on the same items and sampling seed. The sparse columns are separate runs from Table~\ref{tab:quality-main}'s.}
\label{tab:suffix-flush-quality}
\centering
\small
\begin{tabular}{l rrr}
\toprule
& Dense & Ours & Ours, flushed \\
\midrule
$\le$256K (403 items) & 53.1 & 54.8 & 53.8 \\
$>$256K (67 items) & 65.7 & 61.2 & 58.2 \\
All (470) & 54.9 & 55.7 & 54.5 \\
\bottomrule
\end{tabular}
\end{table}

With this policy the decode time stays nearly flat as the generation grows (Figure~\ref{fig:generated-length} and Table~\ref{tab:generated-length-1m}).
Aggregate LongBench-v2 accuracy changes little with flushing (Table~\ref{tab:suffix-flush-quality}).
We did not test reasoning traces as long as 64K tokens.

\Needspace{8\baselineskip}
\section{Design choices on Qwen3.5-4B}
\label{app:design}
\suppressfloats[t]

\subsection{The routing key}
\label{app:routing-space}
We compare routing representations while keeping attention computation unchanged.
The main configuration uses unit-normalized pre-RoPE keys and queries, with spherical $k$-means.
We test using only the 192 non-rotary dimensions, since the model rotates 64 of its 256 key dimensions.
We also test raw keys with Euclidean $k$-means, using unnormalized cluster means and queries at the attention scale $\tau=\sqrt{d}$.

At 512K, both variants stay within about two points of our multi-key accuracy (Table~\ref{tab:routing-space}).
The non-rotary variant also matches our multi-key accuracy at 1M.
Post-RoPE routing gives similar multi-key accuracy to pre-RoPE routing.
PCA projections fitted to raw-corpus keys retain most of the multi-key accuracy at 128 and 32 dimensions.
For timing, we use random projections of the same size.
Reducing the dimension from 256 to 32 shortens the 512K prefill by about 5\% and leaves decoding unchanged.
These results support using the original keys for routing on Qwen3.5.
On Nemotron, however, centering the keys prevents severe cluster imbalance (Table~\ref{tab:nemotron-pilot}).

\begin{table}[t]
\caption{The routing key on Qwen3.5-4B, with RULER multi-key and multi-value accuracy in percent at 512K, $K=L/16$, $u=2{,}048$, 100 items per entry. Our row, the post-RoPE row, and the multi-key entries of the non-rotary row are means over three clustering seeds, and all other entries use seed 0. A dash marks an entry that was not measured.}
\label{tab:routing-space}
\centering
\small
\begin{tabular}{l ccc}
\toprule
Routing key & MK-3 & MK-2 & MV \\
\midrule
Dense & 96.0 & 99.0 & 100.0 \\
Whole key, normalized, $\tau=0.1$ (ours) & 92.7 & 95.0 & 100.0 \\
Whole key after RoPE, normalized, $\tau=0.1$ & 93.0 & 94.0 & 100.0 \\
Non-rotary 192 dimensions, normalized, $\tau=0.1$ & 93.0 & 95.7 & 99.5 \\
Raw key, Euclidean $k$-means, $\tau=16$ & 91.0 & 93.0 & 99.3 \\
Raw key, Euclidean $k$-means, $\tau=4$ & 95.0 & 96.0 & 100.0 \\
PCA to 128 dimensions, normalized, $\tau=0.1$ & 94.0 & 91.0 & 100.0 \\
PCA to 32 dimensions, normalized, $\tau=0.1$ & 90.0 & 96.0 & 100.0 \\
\bottomrule
\end{tabular}
\end{table}

\subsection{Token budget and block size at 512K}
\label{app:operating-point}
Table~\ref{tab:operating-point} lists the values plotted in Figure~\ref{fig:operating-point}.
It crosses four mean block sizes $u$ with four token budgets $K$ and scores each point on MK-2 over the same 100 items at 512K.
The main setting's accuracy is the three-seed mean of the main RULER table, and the other points are single runs of one code version.
Each prefill time is the mean over two prompts.
We use MK-2 rather than MK-3 here because its gap to dense attention is larger, so it separates the settings more clearly.

\begin{table}[t]
\centering
\small
\caption{The values behind Figure~\ref{fig:operating-point}, Qwen3.5-4B at 512K.}
\label{tab:operating-point}
\setlength{\tabcolsep}{4pt}
\begin{tabular}{r rrrr rrrr}
\toprule
& \multicolumn{4}{c}{MK-2 (\%) at $K$} & \multicolumn{4}{c}{Prefill (s) at $K$} \\
\cmidrule(lr){2-5} \cmidrule(lr){6-9}
$u$ & 8K & 16K & 32K & 64K & 8K & 16K & 32K & 64K \\
\midrule
Dense & \multicolumn{4}{c}{99.0} & \multicolumn{4}{c}{40.69} \\
\midrule
4{,}096 & 32.0 & 62.0 & 90.0 & 100.0 & 12.66 & 14.05 & 16.78 & 23.47 \\
2{,}048 & 40.0 & 75.0 & 95.0 & 99.0 & 13.40 & 14.98 & 17.86 & 24.30 \\
1{,}024 & 56.0 & 87.0 & 97.0 & 100.0 & 15.30 & 17.31 & 20.42 & 26.14 \\
512 & 73.0 & 86.0 & 97.0 & 99.0 & 18.97 & 22.21 & 25.59 & 30.05 \\
\bottomrule
\end{tabular}

\end{table}

At a fixed budget, finer blocks are slower.
Their accuracy advantage is large at small budgets and narrows as the budget grows.
The coarse partition with the main budget is both faster and more accurate than the fine partition with half that budget.
Because selection takes whole clusters, their total size exceeds $K$ by 0.4--14\% across the grid and 2.4\% at the main setting, excluding tokens in $\mathcal D_t$.
Accounting for this overshoot does not change the comparison.

\begin{figure}[t]
  \centering
  \includegraphics[width=\linewidth]{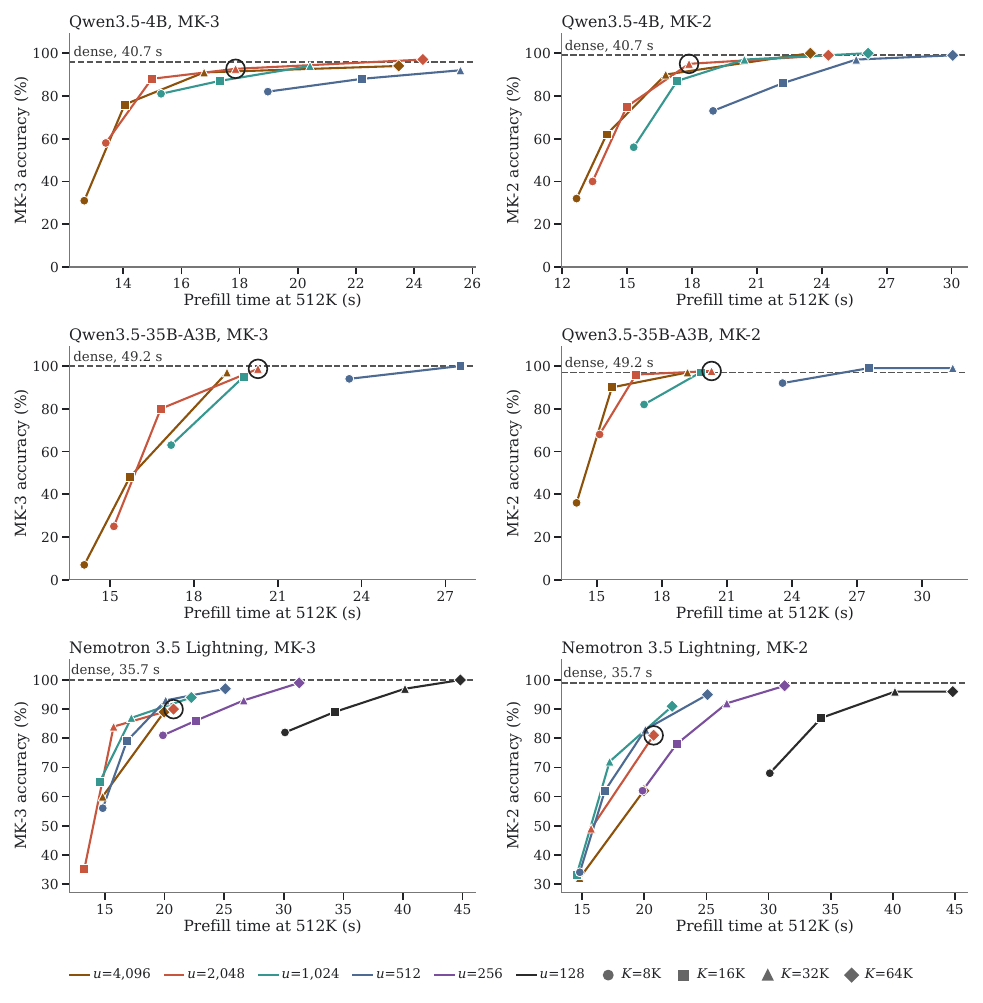}
\caption{Cluster size and token budget on MK-3 and MK-2 at 512K for Qwen3.5-4B, Qwen3.5-35B-A3B, and Nemotron 3.5 Lightning.}
  \label{fig:grid-qwen}
\end{figure}

Figure~\ref{fig:grid-qwen} extends the grid to MK-3 and MK-2 on all three models, with values for the Qwen models in Table~\ref{tab:grid-qwen} and for Nemotron in Table~\ref{tab:nemotron-grid}.
Both Qwen models show the same pattern, with finer clusters ahead at small budgets and coarser clusters catching up as the budget grows.
The main setting's accuracy on each model is the three-seed mean of the main RULER table, and the other points are single runs.
For the 35B-A3B model's $u=512$ timings, gathering for KV-outer attention is split into two batches to fit in device memory.

\begin{table}[t]
\caption{The values behind Figure~\ref{fig:grid-qwen}, Qwen3.5-4B and Qwen3.5-35B-A3B at 512K. A dash marks a point that was not measured.}
\label{tab:grid-qwen}
\centering
\small
\setlength{\tabcolsep}{2.6pt}
\begin{tabular}{r rrrr rrrr rrrr}
\toprule
& \multicolumn{4}{c}{MK-3 (\%) at $K$} & \multicolumn{4}{c}{MK-2 (\%) at $K$} & \multicolumn{4}{c}{Prefill (s) at $K$} \\
\cmidrule(lr){2-5} \cmidrule(lr){6-9} \cmidrule(lr){10-13}
$u$ & 8K & 16K & 32K & 64K & 8K & 16K & 32K & 64K & 8K & 16K & 32K & 64K \\
\midrule
\multicolumn{13}{l}{\textit{Qwen3.5-4B}} \\
Dense & \multicolumn{4}{c}{96.0} & \multicolumn{4}{c}{99.0} & \multicolumn{4}{c}{40.69} \\
4{,}096 & 31.0 & 76.0 & 91.0 & 94.0 & 32.0 & 62.0 & 90.0 & 100.0 & 12.66 & 14.05 & 16.78 & 23.47 \\
2{,}048 & 58.0 & 88.0 & 92.7 & 97.0 & 40.0 & 75.0 & 95.0 & 99.0 & 13.40 & 14.98 & 17.86 & 24.30 \\
1{,}024 & 81.0 & 87.0 & 94.0 & -- & 56.0 & 87.0 & 97.0 & 100.0 & 15.30 & 17.31 & 20.42 & 26.14 \\
512 & 82.0 & 88.0 & 92.0 & -- & 73.0 & 86.0 & 97.0 & 99.0 & 18.97 & 22.21 & 25.59 & 30.05 \\
\midrule
\multicolumn{13}{l}{\textit{Qwen3.5-35B-A3B}} \\
Dense & \multicolumn{4}{c}{100.0} & \multicolumn{4}{c}{97.0} & \multicolumn{4}{c}{49.19} \\
4{,}096 & 7.0 & 48.0 & 97.0 & -- & 36.0 & 90.0 & 97.0 & -- & 14.07 & 15.70 & 19.18 & -- \\
2{,}048 & 25.0 & 80.0 & 98.7 & -- & 68.0 & 96.0 & 97.7 & -- & 15.13 & 16.81 & 20.29 & -- \\
1{,}024 & 63.0 & 95.0 & -- & -- & 82.0 & 97.0 & -- & -- & 17.18 & 19.78 & -- & -- \\
512 & 94.0 & 100.0 & -- & -- & 92.0 & 99.0 & 99.0 & -- & 23.56 & 27.54 & 31.40 & -- \\
\bottomrule
\end{tabular}

\end{table}

\begin{figure}[t]
\centering
\includegraphics[width=0.62\linewidth]{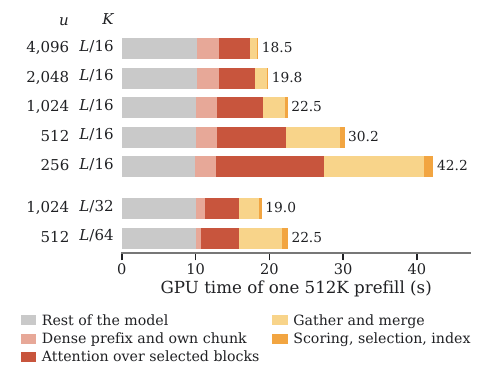}
\caption{GPU time of one 512K prefill on Qwen3.5-35B-A3B by cluster size $u$ and token budget $K$, drawn as Figure~\ref{fig:cluster-size-profile}.}
\label{fig:profile-appendix}
\end{figure}

Figure~\ref{fig:profile-appendix} shows the corresponding profile for Qwen3.5-35B-A3B.
The dense-prefix cutoff scales with the token budget.
In the 4B experiment of Appendix~\ref{app:dense-prefix}, reducing this cutoff did not improve prefill time.

\subsection{Budget scaling across context lengths}
\label{app:budget-scaling}

\begin{table}[t]
\caption{Qwen3.5-4B, MK-3 accuracy in percent ($n=100$) and prefill speedup at each length and budget. The $K=L/16$ column is the three-seed mean of the main table, and the other entries are single runs.}
\label{tab:budget-scaling}
\centering
\small
\begin{tabular}{l c cccc}
\toprule
Prompt & Dense & $K=L/8$ & $K=L/16$ & $K=L/32$ & $K=L/64$ \\
\midrule
256K & 100 {\color{gray}\scriptsize(12.4\,s)} & 97 {\color{gray}\scriptsize(1.36$\times$)} & 97.7 {\color{gray}\scriptsize(1.74$\times$)} & 88 {\color{gray}\scriptsize(1.86$\times$)} & 42 {\color{gray}\scriptsize(1.98$\times$)} \\
512K & 96 {\color{gray}\scriptsize(40.7\,s)} & 95 {\color{gray}\scriptsize(1.62$\times$)} & 92.7 {\color{gray}\scriptsize(2.25$\times$)} & 89 {\color{gray}\scriptsize(2.68$\times$)} & 57 {\color{gray}\scriptsize(2.98$\times$)} \\
1M & 40 {\color{gray}\scriptsize(144.1\,s)} & 32 {\color{gray}\scriptsize(1.83$\times$)} & 28.3 {\color{gray}\scriptsize(2.84$\times$)} & 24 {\color{gray}\scriptsize(3.68$\times$)} & 18 {\color{gray}\scriptsize(4.33$\times$)} \\
\bottomrule
\end{tabular}

\end{table}

Increasing the budget beyond $L/16$ gives little additional MK-3 accuracy while increasing latency (Table~\ref{tab:budget-scaling}).
Reducing the budget lowers latency at the cost of retrieval accuracy.
Neither a fixed token count nor a fixed fraction predicts accuracy across lengths, because the dense model's own retrieval performance also declines.

\begin{figure}[t]
  \centering
  \includegraphics[width=\linewidth]{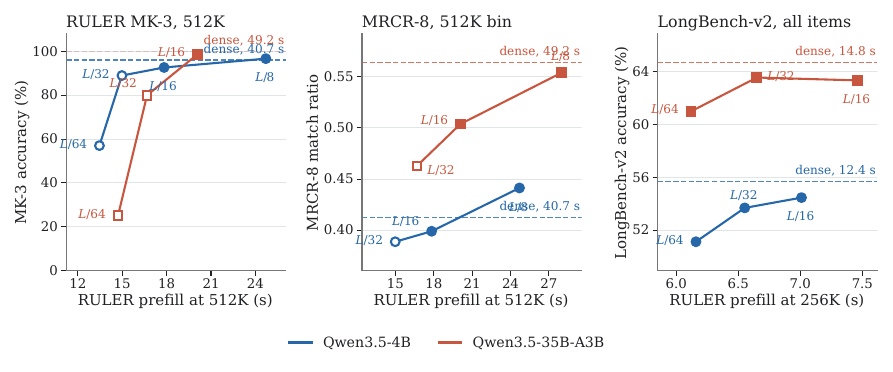}
\caption{Quality against prefill time over token budgets from $L/64$ to $L/8$ on Qwen3.5-4B and Qwen3.5-35B-A3B.}
  \label{fig:budget-quality}
\end{figure}

Figure~\ref{fig:budget-quality} extends the comparison at 512K to MRCR-8 and LongBench-v2 on both Qwen models, over budgets from $L/64$ to $L/8$.
Because benchmark prompt lengths vary, we use the prefill time of a fixed-length RULER prompt as a common reference on the horizontal axis.
For each budget, the reference prompt is 512K for MRCR-8 and 256K for LongBench-v2, the upper bound of the band containing most LongBench-v2 items.
Filled markers are means over three runs, and hollow markers are single runs.

MK-3 accuracy declines below $L/16$, while MRCR-8 changes less.
At $L/32$, the MRCR-8 scores stay within the range of the three $L/16$ runs (Table~\ref{tab:sink-mrcr-longbench}).
Increasing the budget to $L/8$ raises MRCR-8 above this range on the 4B model but not on the 35B-A3B model.
On the 35B-A3B model, 512-token clusters match dense MK-3 accuracy at $L/32$ and come within six points at $L/64$, but take longer to prefill than the main setting (Table~\ref{tab:grid-qwen}).
LongBench-v2 holds its accuracy down to $L/32$ and drops at $L/64$.

\subsection{Positional baselines}
\label{app:positional}

We form contiguous blocks of $B$ tokens and represent each block by its mean pre-RoPE key.
Queries score these means without vector normalization at temperature $\sqrt{d}$, and head probabilities are summed within each GQA group before selecting the top $\lceil K/B\rceil$ blocks.
With the positional baseline's raw keys, unnormalized cluster means, and temperature $\sqrt{d}$, semantic blocks lose only a few points of accuracy (Table~\ref{tab:routing-space}).
They still outperform positional blocks, so normalization and temperature do not explain the gap.
Both methods use the same token budget and dense-prefix, current-chunk, and generated-suffix policies.
With per-query selection, positional blocks use the same KV-outer execution as semantic blocks (Section~\ref{sec:implementation}).

For shared selection, $g$ consecutive queries select blocks together using the tile score of FlashPrefill V2~\citep{fan2026flashprefillv2}.
Within each query head, we sum the exponentiated block scores across the queries and normalize over blocks.
We then sum these probabilities within each GQA group and select the highest-scoring blocks to fill the budget.
All $g$ queries attend to the selected blocks with their original query vectors and causal masks.
Groups stay within a prefill chunk, and an incomplete final group is padded by repeating its last query.
Both block layouts use this procedure, while decoding remains per-query.
A Triton kernel computes group scores without storing individual query scores, and each group attends to its selected blocks in one FlashAttention call~\citep{jiang2024minference,lai2025flexprefill}.

FlashPrefill V2 shares one selection across 128 query rows.
On the 4B model, these correspond to four query heads over 32 positions, within the sharing spans tested here~\citep{fan2026flashprefillv2}.
Our comparisons vary block layout and query sharing at a fixed token budget; they do not reproduce the complete selection policies of MInference, FlexPrefill, QuoKA, or FlashPrefill.
The released MInference and FlexPrefill kernels available for these experiments supported head dimensions up to 128, whereas Qwen3.5 uses 256.
We found no public QuoKA code when we ran the comparisons.
Appendix~\ref{app:flashprefill} compares directly with FlashPrefill V2.

At most settings, we also tested scoring blocks with the group's mean query, as in MInference and FlexPrefill~\citep{jiang2024minference,lai2025flexprefill}.
We further tested the maximum score across queries and a sum over both queries and GQA heads as in FlashPrefill V2.
Each gave accuracy similar to or lower than the summed exponentials.
Table~\ref{tab:positional-points} lists all points in Figure~\ref{fig:positional}, without and with the initial block.

\begin{table}[t]
\caption{Positional and semantic retrieval on Qwen3.5-4B at 512K, $K=L/16$, MK-3 and MK-2 accuracy in percent with $n=100$, without and with the fixed initial block of Appendix~\ref{app:sink}. $B$ is positional block size and $g$ query sharing span. Min/max rows describe each positional block by its minimum and maximum keys, and their shared rows use the averaged query. Our per-token row is the mean over three clustering seeds. A dash denotes an unmeasured entry.}
\label{tab:positional-points}
\centering
\small
\begin{tabular}{l r r r r r r r}
\toprule
 & & & \multicolumn{2}{c}{MK-3 (\%), initial block} & \multicolumn{2}{c}{MK-2 (\%), initial block} & \\
\cmidrule(lr){4-5} \cmidrule(lr){6-7}
Index & $B$ & $g$ & Without & With & Without & With & Prefill (s) \\
\midrule
Positional, shared & 2,048 & 64 & 7 & 4 & -- & -- & 15.6 \\
 & 128 & 64 & 27 & 32 & -- & -- & 16.2 \\
 & 32 & 64 & 51 & 64 & 24 & 24 & 18.4 \\
 & 16 & 64 & 63 & 79 & 32 & 36 & 21.5 \\
 & 8 & 64 & 77 & 92 & 51 & 53 & 27.5 \\
 & 4 & 64 & 75 & 85 & -- & -- & 39.5 \\
 & 32 & 32 & 64 & 78 & 35 & 38 & 19.1 \\
 & 16 & 32 & 82 & 94 & 44 & 52 & 24.0 \\
 & 8 & 32 & 84 & 93 & 58 & 64 & 32.9 \\
 & 4 & 32 & 85 & 93 & 65 & 73 & 50.1 \\
 & 32 & 16 & 78 & 85 & 51 & 54 & 23.8 \\
Positional, per query & 2,048 & 1 & 10 & 9 & -- & -- & 16.1 \\
 & 128 & 1 & 49 & 51 & 34 & 37 & 33.1 \\
 & 32 & 1 & 82 & 89 & 70 & 69 & 84.9 \\
\midrule
Min/max, shared & 2,048 & 64 & 6 & 6 & -- & -- & -- \\
 & 32 & 64 & 24 & 21 & -- & -- & 18.7 \\
 & 16 & 64 & 26 & 25 & -- & -- & -- \\
Min/max, per query & 32 & 1 & 44 & 44 & -- & -- & 87.7 \\
\midrule
Ours, per query & -- & 1 & 92.7 & 95.0 & 95.0 & 94.7 & 18.3 \\
Ours, shared & -- & 64 & 26 & 21 & 11 & 18 & 16.4 \\
Dense & -- & -- & 96 & -- & 99 & -- & 40.7 \\
\bottomrule
\end{tabular}
\end{table}

\begin{table}[t]
\caption{Positional and semantic retrieval on Qwen3.5-35B-A3B at 512K, $K=L/16$, without the initial block, MK-3 and MK-2 accuracy in percent with $n=100$ (ours on clustering seed 0) and 512K prefill time.}
\label{tab:positional-points-35b}
\centering
\small
\begin{tabular}{l r r r r r}
\toprule
Index & $B$ & $g$ & MK-3 (\%) & MK-2 (\%) & Prefill (s) \\
\midrule
Positional, shared & 16 & 64 & 78 & 58 & 30.8 \\
 & 32 & 32 & 84 & 63 & 22.5 \\
 & 16 & 32 & 93 & 66 & 27.4 \\
 & 8 & 32 & 99 & 84 & 37.1 \\
Positional, per query & 128 & 1 & 55 & 47 & 34.4 \\
\midrule
Ours, per query & -- & 1 & 99 & 97 & 20.4 \\
Dense & -- & -- & 100 & 97 & 49.2 \\
\bottomrule
\end{tabular}
\end{table}

On Qwen3.5-35B-A3B (Table~\ref{tab:positional-points-35b}), MK-3 is near its ceiling, and positional blocks of 8 tokens shared over 32 queries reach our accuracy, but their prefill takes $1.8\times$ as long, as on the 4B model.
Every faster positional setting is at least 6 points less accurate.
On MK-2, the same setting is 14 points below our three-seed mean of 97.7\%, and every other positional setting is lower still.

In Table~\ref{tab:positional-points}, the columns with the initial block add the fixed initial block of Appendix~\ref{app:sink} to every row, which changes the prefill times by at most 5\%.

Table~\ref{tab:positional-points-l32} repeats the comparison at the smaller budget $K=L/32$, where our accuracy falls further below dense attention.
Our method stays faster than every positional setting we timed, and is more accurate on MK-2 with and without the initial block.
With the block, the most accurate positional setting approaches our MK-3 accuracy but takes $1.9\times$ as long to prefill as our method at $u=1{,}024$ and $2.2\times$ as long as at $u=2{,}048$.
At this budget, halving our cluster size to $u=1{,}024$ recovers much of the MK-2 loss, as in the grid of Appendix~\ref{app:operating-point}.

\begin{table}[t]
\caption{Positional and semantic retrieval on Qwen3.5-4B at 512K with the smaller budget $K=L/32$, accuracy in percent with $n=100$, without and with the fixed initial block, and 512K prefill time without it.}
\label{tab:positional-points-l32}
\centering
\small
\begin{tabular}{l r r r r r r r}
\toprule
 & & & \multicolumn{2}{c}{MK-3 (\%), initial block} & \multicolumn{2}{c}{MK-2 (\%), initial block} & \\
\cmidrule(lr){4-5} \cmidrule(lr){6-7}
Index & $B$ & $g$ & Without & With & Without & With & Prefill (s) \\
\midrule
Positional, shared & 16 & 32 & 71 & 83 & 32 & 36 & 22.1 \\
 & 8 & 64 & 61 & 81 & 39 & 48 & 26.8 \\
 & 8 & 32 & 72 & 88 & 49 & 60 & 32.0 \\
 & 4 & 32 & 64 & 79 & 58 & 62 & 51.5 \\
Positional, per query & 32 & 1 & 68 & 69 & 47 & 50 & -- \\
\midrule
Ours, per query, $u=2{,}048$ & -- & 1 & 82.7 & 90.0 & 76.7 & 86.0 & 14.8 \\
Ours, per query, $u=1{,}024$ & -- & 1 & 87.0 & 90.7 & 88.0 & 89.3 & 17.1 \\
Dense & -- & -- & 96 & -- & 99 & -- & 40.1 \\
\bottomrule
\end{tabular}
\end{table}

A Quest-inspired min/max descriptor does not improve the positional baseline in this pre-RoPE setting~\citep{tang2024quest}.
This comparison tests the descriptor, not Quest's complete decoding method.
Its shared rows and the temperature sweeps below score the group's averaged query rather than the summed exponentials.

The min/max descriptor bounds the dot product with each block's pre-RoPE keys.
It is not an upper bound on the model's post-RoPE attention logits.
At $B=32$ with shared selection, increasing its scoring temperature from 16 to 64, 256, or 1,024 gives MK-3 scores of 19, 20, and 19 percent, respectively.
The mean-key descriptor's temperature, swept at the same point over 4, 16, 64, 256 and 1,024, gives 46, 51, 49, 44 and 41 percent, so the baseline's default temperature, $\sqrt{d}=16$, is the best of the five, and none moves the baseline towards semantic blocks.

\subsection{Comparison with FlashPrefill V2}
\label{app:flashprefill}
FlashPrefill V2 is the only practical training-free sparse prefill method we found whose released kernels support the 256-dimensional heads of Qwen3.5 on H200~\citep{fan2026flashprefillv2}.
We run it with the SGLang implementation released with it, on Qwen3.5-4B with the same 512K prompts and YaRN setting as our tables.
It uses the settings of the launch example in its repository, except that all eight full-attention layers are sparse, as in our method.
We vary its threshold $\alpha$, which sets how many blocks it keeps. It always keeps the first blocks of the prompt as attention sinks.
Both methods run in BF16.
SGLang times are the median time to the first token over the 100 prompts, with prefill in chunks of 16,384 tokens, and our times are the prefill time of the latency prompt (Appendix~\ref{app:protocol}).
FlashPrefill V2 is one run per threshold. Our points by budget are single runs timed in one session, as in Appendix~\ref{app:operating-point}, and the point with the initial block is the mean over three clustering seeds.

\begin{figure}[t]
\centering
\includegraphics[width=0.85\linewidth]{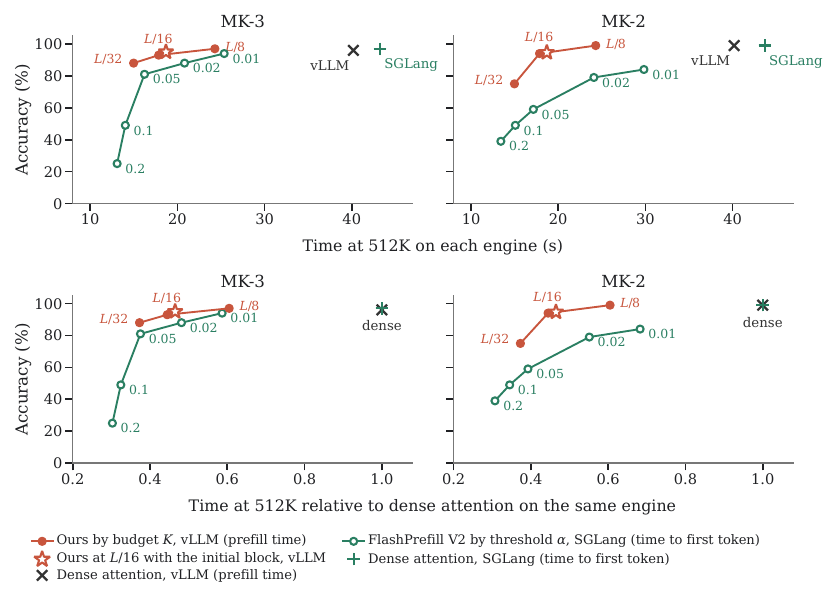}
\caption{FlashPrefill V2 in SGLang and our method in vLLM on Qwen3.5-4B at 512K, against time on each engine (top) and against time relative to dense attention on the same engine (bottom). Labels are our budget $K$ and the threshold $\alpha$ of FlashPrefill V2.}
\label{fig:flashprefill}
\end{figure}

Figure~\ref{fig:flashprefill} and Table~\ref{tab:flashprefill} report accuracy and latency on each engine.
The dense baselines have similar latency and accuracy, but the engines and timing protocols differ.
We therefore compare each method's speedup over its own engine's dense baseline.
Our method is more accurate at comparable speedups on both tasks.
Among the tested thresholds, FlashPrefill V2 reaches our MK-3 accuracy only with a $1.7\times$ speedup, compared with $2.2\times$ for our method at $K=L/16$.

\begin{table}[t]
\caption{FlashPrefill V2 in SGLang and our method in vLLM on Qwen3.5-4B at 512K. Accuracy is on 100 prompts per task. Times are the prefill time of the latency prompt on vLLM and the median time to the first token over the 100 prompts on SGLang.}
\label{tab:flashprefill}
\centering
\small
\begin{tabular}{ll cc cc}
\toprule
 & & \multicolumn{2}{c}{Accuracy (\%)} & \multicolumn{2}{c}{Time at 512K (s)} \\
\cmidrule(lr){3-4} \cmidrule(lr){5-6}
Method & Engine & MK-3 & MK-2 & MK-3 run & MK-2 run \\
\midrule
Dense attention & vLLM & 96.0 & 99.0 & \multicolumn{2}{c}{40.18} \\
Ours, $K=L/32$ & vLLM & 88.0 & 75.0 & \multicolumn{2}{c}{14.98} \\
Ours, $K=L/16$ & vLLM & 93.0 & 94.0 & \multicolumn{2}{c}{17.86} \\
Ours, $K=L/8$ & vLLM & 97.0 & 99.0 & \multicolumn{2}{c}{24.30} \\
Ours, $K=L/16$, initial block & vLLM & 95.0 & 94.7 & \multicolumn{2}{c}{18.69} \\
\midrule
Dense attention & SGLang & 97.0 & 99.0 & 43.25 & 43.70 \\
FlashPrefill V2, $\alpha=0.2$ & SGLang & 25.0 & 39.0 & 13.09 & 13.42 \\
FlashPrefill V2, $\alpha=0.1$ & SGLang & 49.0 & 49.0 & 14.03 & 15.08 \\
FlashPrefill V2, $\alpha=0.05$ & SGLang & 81.0 & 59.0 & 16.23 & 17.15 \\
FlashPrefill V2, $\alpha=0.02$ & SGLang & 88.0 & 79.0 & 20.82 & 24.09 \\
FlashPrefill V2, $\alpha=0.01$ & SGLang & 94.0 & 84.0 & 25.37 & 29.83 \\
\bottomrule
\end{tabular}

\end{table}

\subsection{Centroids fitted on a corpus}
\label{app:routing-variants}
\label{sec:results-ablation}

Table~\ref{tab:centroid-source} compares centroids fitted to each prompt with centroids fitted once and reused, with 256 centroids per layer and KV head.
All rows keep the original query and key representations and the same selection rule, so this comparison does not include SAAP's learned query classifier~\citep{mazare2025saap}.
Per-prompt results average three clustering seeds, and the other rows are single runs.
Raw-corpus centroids severely reduce multi-key accuracy and assign over half of the common-word prompt to one cluster, which the runtime rejects, so the table marks that task as failed.
Fitting once to twenty RULER-5 examples per task recovers much of the loss.
This fitting set includes evaluation examples, so the row diagnoses corpus mismatch rather than held-out generalization.
The first-64K row fits each prompt's centroids on its first 64K tokens and assigns every later key to them, as the prefix index of Appendix~\ref{app:memory} does.

\clearpage
\section{Sources of quality loss and training baselines}
\label{app:loss}
\suppressfloats[t]

\subsection{Attention mass missed by cluster scoring}
\label{app:mass-decomposition}

Cluster selection can miss high-attention tokens because centroid scores misrank clusters or because whole clusters also contain low-attention tokens.
We examine these effects by comparing centroid-based selection with dense-attention rankings, using the same partition.

We build the partition from the dense model's keys and compare three selections for the dense model's queries.
The first applies our selection rule to the fitted clusters.
The second sums the dense attention probabilities within each cluster and greedily selects whole clusters under the same budget, ranking by total mass or by mass per token and keeping the better of the two for each query and GQA group.
The third selects the $K$ tokens that receive the most dense attention, without reading whole clusters.
We call the last two \emph{dense-ranked clusters} and \emph{dense-ranked tokens}.
Both need dense attention at inference time, so they serve only as references.

Coverage is the percentage of dense attention to the prompt retained by the selected tokens.
Each query's current chunk is always read exactly and does not count toward the budget.
For generated queries, we count only attention to prompt tokens.

The analysis uses ten prompts per task at 512K, with the last 64 prompt positions and up to 32 generated tokens as queries, over the 4B model's eight full-attention layers and four KV heads.
The ten MK-3 prompts come from a separate set of 400 generated prompts rather than from the MK-3 items of our tables, and MRCR uses the first ten prompts of its 512K bin.
The budgets are $L/16$ and $L/8$ of each prompt's actual length.
Dense probabilities are summed over a GQA group to rank tokens or clusters, and the retained mass is normalized by each query head's attention on the prompt and averaged over heads, queries, layers, and prompts.
Because heads can put different total attention on the prompt during generation, the ranking and the reported coverage are not exactly the same quantity.
These queries emphasize retrieval near the end of the prompt and during generation, not the average prefill query.

\begin{table}[t]
\centering
\small
\caption{Mean prompt-attention coverage on ten 512K examples per task, Qwen3.5-4B.}
\label{tab:mass-decomposition}
\setlength{\tabcolsep}{5pt}
\begin{tabular}{lc rrr}
\toprule
& & \multicolumn{3}{c}{Prompt attention retained (\%)} \\
\cmidrule(lr){3-5}
Task & $K$ & Our scores & Dense-ranked clusters & Dense-ranked tokens \\
\midrule
MK-2 & $L/16$ & 77.4 & 80.8 & 90.4 \\
MK-2 & $L/8$ & 84.6 & 87.3 & 94.3 \\
\addlinespace[3pt]
MK-3 & $L/16$ & 76.9 & 80.5 & 91.9 \\
MK-3 & $L/8$ & 84.0 & 86.8 & 95.1 \\
\addlinespace[3pt]
MRCR-8 & $L/16$ & 86.3 & 88.2 & 94.2 \\
MRCR-8 & $L/8$ & 91.0 & 92.2 & 96.6 \\
\bottomrule
\end{tabular}

\end{table}

\begin{figure}[t]
\centering
\includegraphics[width=\linewidth]{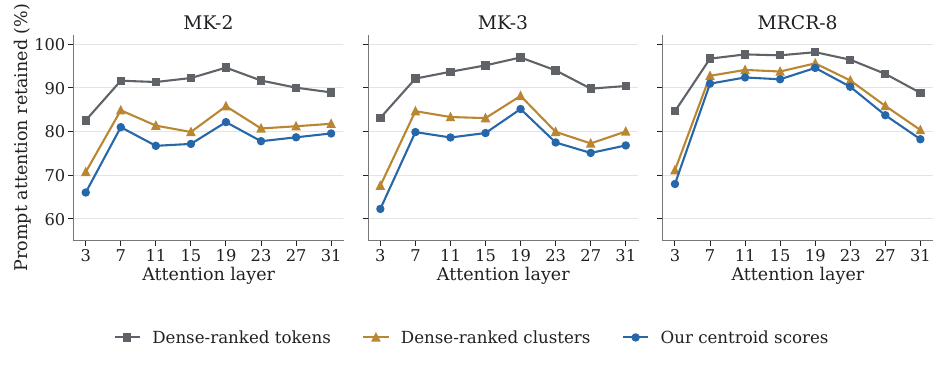}\\[2pt]
\includegraphics[width=\linewidth]{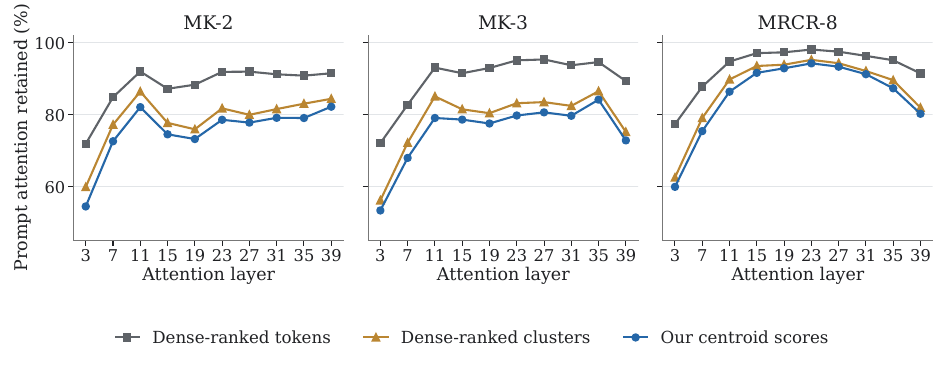}
\caption{Prompt-attention coverage by full-attention layer at $K=L/16$ under our centroid scores, dense-ranked clusters, and dense-ranked tokens, on ten 512K examples per task, for Qwen3.5-4B (top, eight layers) and Qwen3.5-35B-A3B (bottom, ten). All curves use dense-model states.}
\label{fig:mass-decomposition}
\end{figure}

At $K=L/16$, dense-ranked tokens retain more attention mass than dense-ranked clusters on all three tasks (Figure~\ref{fig:mass-decomposition}).
Cluster selections can exceed the budget by part of a cluster, but giving the token reference the same number of tokens changes its coverage by at most 0.3 points.
Better cluster scores therefore recover only part of the mass lost when selection is restricted to whole clusters.
Using our centroid scores with $K=L/8$ also gives higher coverage than dense-ranked clusters with $K=L/16$ on all three tasks (Table~\ref{tab:mass-decomposition}).

We also replace the centroid scores inside the sparse runtime and measure RULER accuracy (Table~\ref{tab:cluster-oracle}).
At every full-attention layer, we compute dense attention probabilities from the current queries and prompt keys in FP32, sum them within each cluster, and select clusters in that order under the runtime's budget rule.
The clustering, the regions read without selection, the attention computation, and the handling of generated tokens are unchanged.
Queries and keys come from this sparse run, so the comparison includes the effect of earlier sparse layers on the model states.
We use the first 50 prompts of each multi-key task at 512K, with clustering seed 0 for the dense-attention scores and the mean and standard deviation over three seeds for the centroid scores.

\begin{table}[t]
\centering
\small
\caption{Cluster scoring on Qwen3.5-4B at 512K, $u=2{,}048$, $K=L/16$.
Accuracy (\%) on 50 examples per task.}
\label{tab:cluster-oracle}
\setlength{\tabcolsep}{4pt}
\begin{tabular}{l cc}
\toprule
Selection & MK-3 & MK-2 \\
\midrule
Dense attention & 96.0 & 98.0 \\
Centroid scores & 93.3 (1.2) & 96.0 (0.0) \\
Dense attention mass & 94.0 & 98.0 \\
\bottomrule
\end{tabular}
\end{table}

Using dense-attention scores recovers dense accuracy on MK-2 but changes MK-3 little (Table~\ref{tab:cluster-oracle}).
The effect of missing attention mass depends on which tokens are missed.
A token carrying answer information matters more than one repeating information available elsewhere.

\subsection{Learning the routing map}
\label{app:learned-projection}
We test two simple training baselines.
The first learns linear maps for selection while leaving the model weights and attention kernels unchanged.
For each full-attention layer, we train a query map $W_q\in\mathbb R^{d\times d}$ per query head and a key map $W_k\in\mathbb R^{d\times d}$ per KV head, initialized to the identity.
We also train the temperature $\tau$ and size penalty $\beta$.
The penalty is clamped to nonnegative values in the forward pass with a straight-through gradient, matching its use at inference time.

For each training document, we fit clusters on the mapped keys and sum the dense model's attention probabilities within each cluster.
The loss is the KL divergence from this cluster-mass distribution to the normalized selection scores in Equation~\ref{eq:gqa-score}, with an L2 penalty keeping the maps near the identity.
Cluster assignments are recomputed without gradients.
The normalized means of the assigned mapped keys carry gradients to $W_k$.

Training uses 25,684 raw-text documents of 65,536 tokens each, about 1.7B tokens, drawn from the PDF documents of the Dolma~3 Longmino mix~\citep{teamolmo2025olmo3}. We leave out the mix's synthetic common-word extraction and rewriting tasks, which resemble RULER tasks.
We train for 3,000 steps with eight data-parallel workers, a total batch size of eight documents, and learning rate $10^{-3}$.
The objective is evaluated on eight held-out documents and largely plateaus after roughly half the run.
For downstream evaluation, RULER scores are percentages averaged over three clustering seeds, with standard deviations in parentheses.
MRCR-8 reports the mean match ratio over 100 examples.

\begin{table}[t]
\caption{Learning query and key maps for clustering and selection on Qwen3.5-4B.
The left columns report held-out KL loss, retained attention mass, and selection parameters.
The right columns report long-context task performance.}
\label{tab:learned-projection}
\centering
\small
\setlength{\tabcolsep}{3pt}
\resizebox{\linewidth}{!}{%
\begin{tabular}{l cc cc cc cc c}
\toprule
& \multicolumn{2}{c}{Objective} & \multicolumn{2}{c}{Rule} & \multicolumn{2}{c}{512K} & \multicolumn{2}{c}{RULER-5} & MRCR-8 \\
\cmidrule(lr){2-3} \cmidrule(lr){4-5} \cmidrule(lr){6-7} \cmidrule(lr){8-9} \cmidrule(lr){10-10}
Configuration & KL & mass at $L/16$ & $\tau$ & $\beta$ & MK-3 & MK-2 & 256K & 512K & 512K \\
\midrule
Dense attention & -- & -- & -- & -- & 96.0 & 99.0 & 84.7 & 82.8 & 0.381 \\
Identity map (ours) & 1.07 & 0.674 & 0.10 & 0.25 & 92.7 (0.6) & 95.0 (1.0) & 82.7 (0.1) & 81.8 (0.5) & 0.405 \\
Learned $W_q, W_k, \tau, \beta$ & 0.31 & 0.695 & 0.036 & 0 & 91.0 (1.0) & 94.7 (0.6) & 83.1 (0.2) & 81.5 (0.4) & 0.402 \\
\bottomrule
\end{tabular}}
\end{table}

Training reduces the held-out KL by about two thirds, but changes the attention mass retained by the selected clusters much less (Table~\ref{tab:learned-projection}).
Much of the KL reduction accompanies a lower temperature, which sharpens the score distribution, while the size penalty reaches zero.
The objective favors clusters receiving more total attention mass, which can reward large clusters and conflict with the runtime's size penalty.
Allowing a negative penalty reaches a similar objective.

The learned map gives similar aggregate RULER-5 and MRCR-8 scores to the identity map, but its 512K MK-3 accuracy is about two points lower.
Earlier runs used a corpus 8.5 times smaller and a quarter to a third of the training steps, and tested an unconstrained penalty, the same map with the penalty clamped at evaluation, and a query-only map.
None gave a consistent retrieval improvement.

\subsection{Adapting the model to its sparse attention with LoRA}
\label{app:lora-student}
The second baseline adapts the model weights to the sparse attention.
A copy of the model with rank-8 LoRA adapters~\citep{hu2022lora} runs sparse attention and is trained to match the dense model's next-token distribution by forward KL over the full vocabulary.
One configuration adapts the query, key, and output projections of the eight full-attention layers.
The other also adapts the MLP and Gated DeltaNet projections.
The index and the selection are recomputed from the adapted model's current keys and queries with the same code as at inference.
Gradients pass through attention over the selected tokens, but not through the discrete selection.

Each run uses 300 steps with four documents per step, drawn from the corpus of the learned map above.
The documents are 64K tokens long, and the loss is computed at every position.
As an implementation check, the model with zero-initialized adapters reproduces the dense model when both use dense attention.
For evaluation, adapters are merged into the weights and the sparse runtime is unchanged.
Each task uses 100 examples and one clustering seed.

\begin{table}[t]
\caption{Adapting Qwen3.5-4B to sparse attention with LoRA trained on 64K documents.
RULER multi-key scores are accuracy (\%), and MRCR-8 reports a mean match ratio.
All configurations use the same sparse runtime and token budget.}
\label{tab:lora-student}
\centering
\small
\setlength{\tabcolsep}{5pt}
\begin{tabular}{l cc ccc}
\toprule
& \multicolumn{2}{c}{128K} & \multicolumn{3}{c}{512K} \\
\cmidrule(lr){2-3} \cmidrule(lr){4-6}
Adapters & MK-2 & MK-3 & MK-2 & MK-3 & MRCR-8 \\
\midrule
Dense & 100 & 100 & 99 & 96 & 0.381 \\
Ours, no training & 98 & 100 & 94 & 93 & 0.405 \\
Attention & 99 & 99 & 92 & 91 & 0.298 \\
+MLP+GDN & 100 & 99 & 95 & 90 & 0.291 \\
\bottomrule
\end{tabular}
\end{table}

The held-out distillation objective falls substantially during training, but the task results do not improve (Table~\ref{tab:lora-student}).
At 512K, MK-2 stays close to the training-free configuration, while MK-3 and MRCR-8 are lower.
These runs use raw text and documents shorter than the evaluation prompts.
Other work on adapting models to sparse attention also trains on synthetic long-context tasks~\citep{zhou2026rtpurbo}, which we do not test.

\Needspace{0.3\textheight}
\section{Other architectures}
\label{app:architectures}
\label{app:nemotron}
\suppressfloats[t]

This section provides the full results for Qwen3.5-35B-A3B and Nemotron 3.5 Lightning.

\subsection{Qwen3.5-35B-A3B}
\label{app:qwen35b}
Qwen3.5-35B-A3B has 40 layers, including ten full-attention layers with 16 query heads, two KV heads, and head dimension 256.
It uses 256 experts with eight active and shares the 4B model's tokenizer.
The sparse implementation uses the same query and key representations, clustering settings, and selection rule as the 4B model.
All sparse quality measurements use sliced prefill.
Latency measurements use it from 512K onward, where single-pass prefill exceeds device memory.
The dense baseline uses chunked prefill.

\begin{figure}[t]
  \centering
  \includegraphics[width=0.85\linewidth]{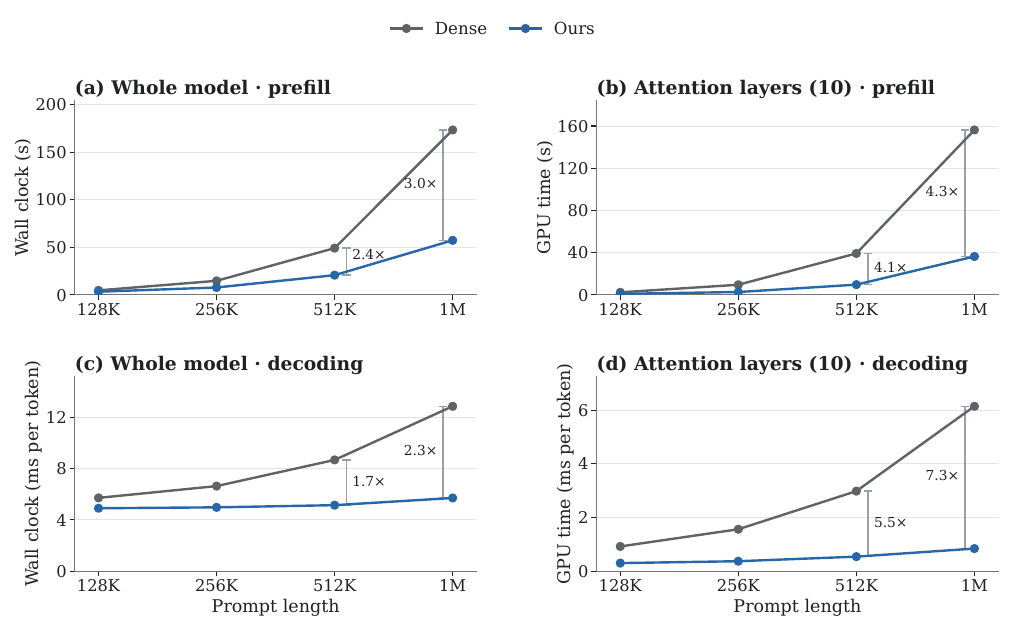}
\caption{Qwen3.5-35B-A3B latency on one H200 at $K=L/16$. The columns show whole-model wall-clock latency (left) and GPU time over the ten full-attention layers (right). The rows show prefill (top) and decoding (bottom).}
  \label{fig:qwen35b-latency}
\end{figure}

Table~\ref{tab:ruler-by-task} gives the task scores beside the 4B model's.
Figure~\ref{fig:qwen35b-latency} shows prefill and decode latency, with values in Table~\ref{tab:qwen35b}.

RULER-5 remains close to dense attention through 1M.
Both multi-key tasks also remain close through 512K.
At 1M, our method loses accuracy on MK-3 and common-word extraction but is above dense attention on MK-2 and HotpotQA.
Halving or quartering the budget at 512K substantially reduces multi-key accuracy.
The QA and common-word results of those runs used capped prompts and are not shown.

\begin{figure}[t]
  \centering
  \begin{minipage}[t]{0.49\linewidth}\centering\includegraphics[width=\linewidth]{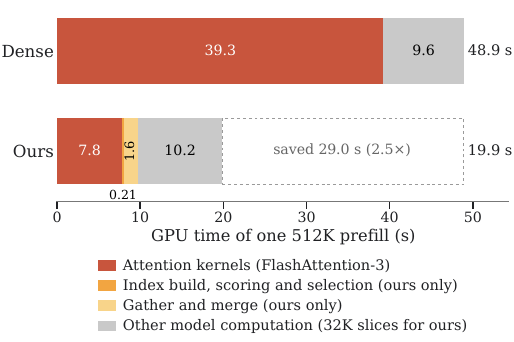}\end{minipage}\hfill
  \begin{minipage}[t]{0.49\linewidth}\centering\includegraphics[width=\linewidth]{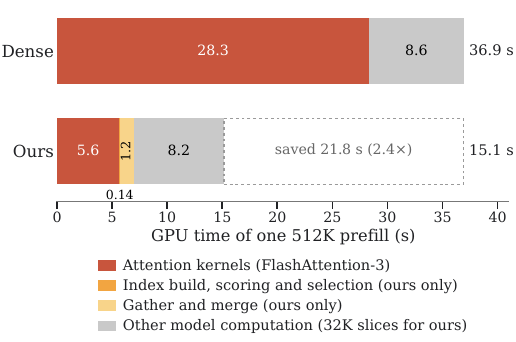}\end{minipage}
\caption{The 512K prefill profile of Qwen3.5-35B-A3B at $K=L/16$ (left) and of Nemotron 3.5 Lightning at $u=2{,}048$ (right), drawn as Figure~\ref{fig:breakdown}, with GPU seconds by kernel group from one profiled prefill of each configuration. Nemotron uses different slicing schedules for non-attention layers in the dense and sparse runs.}
  \label{fig:qwen35b-breakdown}
\end{figure}

The 512K prefill profile in Figure~\ref{fig:qwen35b-breakdown} shows that the unchanged expert and Gated DeltaNet layers account for much of the remaining sparse prefill time.
Because the sparse prefill is sliced at this length, these layers run in 32K slices, and the mixture-of-experts kernels are the largest part of the time not attributed to attention.

For long generations, reading the entire generated suffix exactly increases decode cost.
After a 512K prompt, the decode step grows from 4.9 ms per token at the start to 10.5 ms between 32K and 64K generated tokens, above dense attention's 9.1 ms, while flushing keeps it near 5.1 ms.
Flushing avoids this growth after both 512K and 1M prompts, with similar aggregate LongBench-v2 accuracy in the measured runs (Tables~\ref{tab:qwen35b-generation} and~\ref{tab:qwen35b-k32}).

\begin{table}[t]
\caption{Qwen3.5-35B-A3B latency on one H200 at $K=L/16$ and $u=2{,}048$, as Table~\ref{tab:runtime}. Whole-model columns are wall clock, the attention-layer columns profiler totals over the ten full-attention layers. The last rows halve and quarter the budget at 512K.}
\label{tab:qwen35b}
\centering
\small
\setlength{\tabcolsep}{4pt}
\begin{tabular}{l l cc cc}
\toprule
& & \multicolumn{2}{c}{Whole model} & \multicolumn{2}{c}{Attention layers (10)} \\
\cmidrule(lr){3-4} \cmidrule(lr){5-6}
Prompt & Configuration & Prefill (s) & Decode (ms) & Prefill (s) & Decode (ms) \\
\midrule
128K & Dense & 4.75 & 5.72 & 2.41 & 0.92 \\
128K & Ours & 3.31 & 4.91 & 0.96 & 0.30 \\
256K & Dense & 14.77 & 6.64 & 9.58 & 1.56 \\
256K & Ours & 7.76 & 4.98 & 2.69 & 0.37 \\
512K & Dense & 49.19 & 8.68 & 39.25 & 2.98 \\
512K & Ours & 20.73 & 5.15 & 9.68 & 0.54 \\
1M & Dense & 173.49 & 12.87 & 156.24 & 6.14 \\
1M & Ours & 57.42 & 5.71 & 36.38 & 0.84 \\
\midrule
512K & Ours, $K=L/32$ & 17.21 & 5.03 & -- & -- \\
512K & Ours, $K=L/64$ & 15.21 & 4.96 & -- & -- \\
\bottomrule
\end{tabular}
\end{table}

\begin{table}[t]
\caption{Qwen3.5-35B-A3B, mean decode time after a 1M prompt in milliseconds per token, over 64 and over 65,536 generated tokens, $K=L/16$, one H200.}
\label{tab:qwen35b-generation}
\centering
\small
\begin{tabular}{l cc}
\toprule
Configuration & 64 & 65,536 \\
\midrule
Dense & 12.90 & 13.42 \\
Ours & 5.57 & 9.21 \\
Ours, flushed every 1K ($W=512$) & 5.55 & 5.49 \\
Ours, flushed every 8K ($W=2{,}048$) & 5.57 & 5.91 \\
\bottomrule
\end{tabular}
\end{table}

Table~\ref{tab:qwen35b-k32} compares the main budget with half that budget on MRCR-8 and LongBench-v2 in single runs.
MRCR-8 at the main budget stays close to dense attention through 256K and falls below it at 512K.
Halving the budget lowers MRCR-8 most at 256K.
LongBench-v2 scores change less across the sparse configurations, and flushing changes aggregate accuracy by less than two percentage points.

\begin{table}[t]
\caption{Qwen3.5-35B-A3B on MRCR-8 (mean match ratio) and LongBench-v2 (accuracy over the same 470 items), thinking on, at $K=L/16$ and at half of it. Flushing uses $F=1{,}024$ and $W=512$ (Appendix~\ref{app:suffix-flush}).}
\label{tab:qwen35b-k32}
\centering
\small
\begin{tabular}{l ccc ccc}
\toprule
& \multicolumn{3}{c}{MRCR-8} & \multicolumn{3}{c}{LongBench-v2} \\
\cmidrule(lr){2-4} \cmidrule(lr){5-7}
& 128K & 256K & 512K & $\le$256K & $>$256K & all \\
\midrule
Dense & 0.849 & 0.644 & 0.500 & 65.3 & 65.7 & 65.3 \\
Ours, $K=L/32$ & 0.835 & 0.596 & 0.463 & 64.3 & 65.7 & 64.5 \\
Ours, $K=L/16$ & 0.839 & 0.682 & 0.437 & 62.3 & 64.2 & 62.6 \\
Ours, $K=L/16$, flushed & -- & -- & -- & 63.0 & 71.6 & 64.3 \\
\bottomrule
\end{tabular}
\end{table}

The coverage analysis shows the same ordering as on the 4B model (Figure~\ref{fig:mass-decomposition}).
Mean coverage is slightly lower on the 35B-A3B model even though its task-accuracy loss is smaller.

\subsection{Nemotron 3.5 Lightning}
\label{app:nemotron-section}
Nemotron 3.5 Lightning 30B-A3B combines Mamba-2 and MoE layers with six full-attention layers, each with two KV heads and no rotary embedding.
We replace these attention layers and retain the original model weights.
We also slice non-attention layers to avoid an offset overflow in the Mamba-2 convolution kernel.

For routing, we center keys and use 32 clustering iterations with at least 128 sampled keys per centroid.
These settings were selected on a 64K pilot and retained for longer evaluations (Table~\ref{tab:nemotron-pilot}).
Centering addresses a strong shared direction in the keys that otherwise leads to cluster collapse.
It changes only routing representations, not the KV used for attention.

\begin{table}[t]
\caption{Nemotron 3.5 Lightning, the 64K pilot behind the clustering settings, with MK-3 accuracy in percent on 20 items, $K=L/16$. Centering subtracts the mean key before routing, whitening scales the centered keys by their covariance, and iterations and samples are the $k$-means fit.}
\label{tab:nemotron-pilot}
\centering
\small
\begin{tabular}{l c}
\toprule
Routing representation & MK-3 \\
\midrule
Dense & 100 \\
Uncentered keys & 20 \\
Centered, 32 iterations, 128 samples per centroid & 100 \\
Whitened & 40 \\
Whitened, $k$-means++ seeding, 32 iterations & 35 \\
\bottomrule
\end{tabular}
\end{table}

\begin{table}[t]
\caption{Nemotron 3.5 Lightning latency on one H200 and MK-3 accuracy in percent ($n=100$) by prompt length, at $K=L/16$ unless noted. The attention-layer columns are profiler totals over the six full-attention layers.}
\label{tab:nemotron}
\centering
\small
\setlength{\tabcolsep}{4pt}
\begin{tabular}{l l c cc cc}
\toprule
& & & \multicolumn{2}{c}{Whole model} & \multicolumn{2}{c}{Attention layers (6)} \\
\cmidrule(lr){4-5} \cmidrule(lr){6-7}
Prompt & Configuration & MK-3 (\%) & Prefill (s) & Decode (ms) & Prefill (s) & Decode (ms) \\
\midrule
256K & Dense & 100 & 11.07 & 3.92 & 6.91 & 0.51 \\
256K & Ours, $u=2{,}048$ & 73 & 6.15 & 3.39 & 1.94 & 0.17 \\
256K & Ours, $u=1{,}024$ & 93 & 6.82 & 3.38 & 2.34 & 0.18 \\
256K & Ours, $u=512$ & 98 & 7.68 & 3.42 & 3.07 & 0.22 \\
256K & Ours, $u=2{,}048$, $K=L/8$ & 96 & 7.59 & 3.44 & 3.15 & 0.22 \\
\midrule
512K & Dense & 100 & 35.72 & 4.44 & 28.32 & 0.88 \\
512K & Ours, $u=2{,}048$ & 84 & 15.33 & 3.59 & 6.95 & 0.23 \\
512K & Ours, $u=1{,}024$ & 87 & 17.30 & 3.54 & 8.47 & 0.27 \\
512K & Ours, $u=512$ & 93 & 21.42 & 3.60 & 11.15 & 0.34 \\
512K & Ours, $u=2{,}048$, $K=L/8$ & 90 & 20.70 & 3.63 & 12.06 & 0.32 \\
\midrule
1M & Dense & 100 & 120.18 & 5.37 & 104.12 & 1.64 \\
1M & Ours, $u=2{,}048$ & 70 & 41.09 & 3.65 & 24.54 & 0.35 \\
1M & Ours, $u=2{,}048$, $K=L/8$ & 85 & 61.62 & 3.87 & 44.74 & 0.51 \\
\bottomrule
\end{tabular}

\end{table}

\begin{figure}[t]
  \centering
  \includegraphics[width=0.85\linewidth]{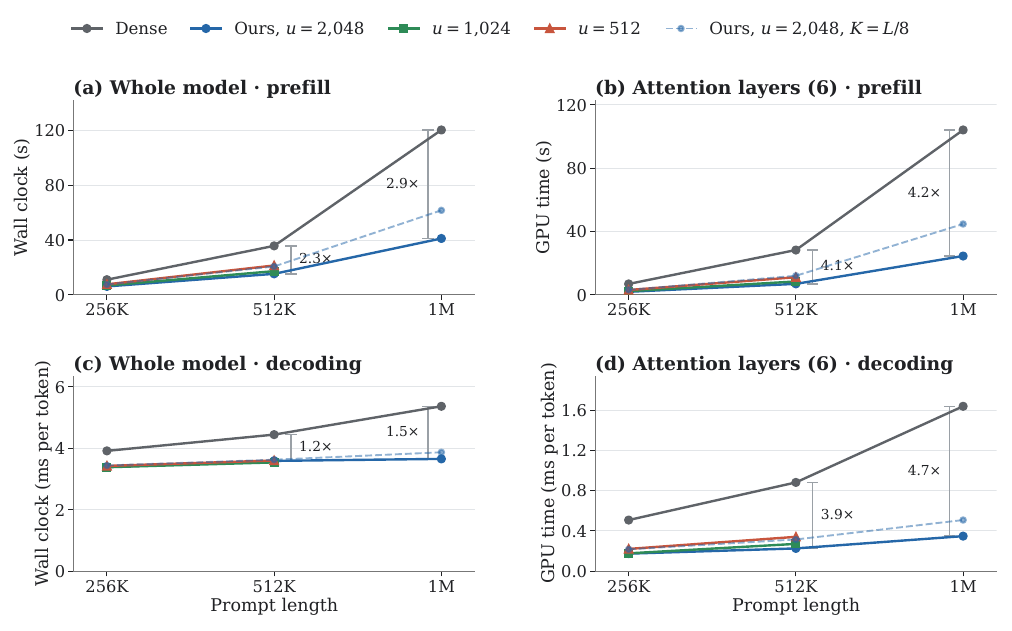}
\caption{Nemotron 3.5 Lightning, latency against prompt length on one H200 at $K=L/16$ for three target mean block sizes $u$, with the doubled budget $K=L/8$ at $u=2{,}048$ dashed. The left column is the whole model by wall clock and the right column the six full-attention layers by profiler accounting. The rows are prefill and decoding. The brackets are dense over ours at $u=2{,}048$ and $K=L/16$, under the conditions of Table~\ref{tab:nemotron}.}
  \label{fig:nemotron-latency}
\end{figure}

Figure~\ref{fig:nemotron-latency} draws the latency columns of Table~\ref{tab:nemotron}.
Both decode paths use full-step graphs.
The nominal 1M latency prompt is 1,028,096 tokens, while the 1M task evaluations use the 1,048,576-token RULER prompts.

At $u=2{,}048$ and $K=L/16$, prefill is $1.8\times$ faster than dense attention at 256K and approaches three times as fast at 1M.
The MK-3 loss is substantial at every length, however.
Decode gains are smaller because the replaced attention layers account for less of the original decode step.

Figure~\ref{fig:grid-qwen} (bottom row) and Table~\ref{tab:nemotron-grid} show the MK-3 and MK-2 grid at 512K, including $u=256$ and 128.
Points below 30\% accuracy are not drawn, and the ringed point is the main setting, $u=2{,}048$ and $K=65{,}536$.
For $u=256$ and 128, KV-outer gathering uses batches of 512 clusters to fit in device memory.

At a fixed budget, accuracy improves as cluster size decreases to 512 tokens.
At similar prefill times, however, 512- or 1,024-token clusters with a larger budget are more accurate than 256- or 128-token clusters.
Nemotron therefore benefits from finer clusters than the Qwen models.
The main results keep $u=2{,}048$, as on the Qwen models.
At $K=L/8$, changing to $u=1{,}024$ raises MK-2 accuracy by 10 points for about 7\% more prefill time.

\begin{figure}[t]
\centering
\includegraphics[width=0.62\linewidth]{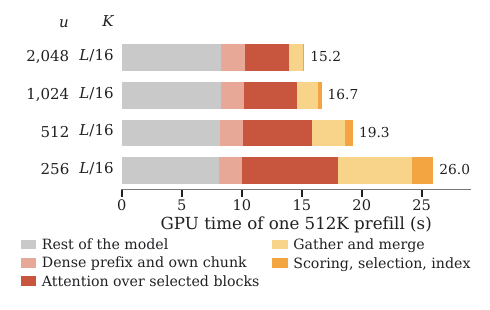}
\caption{GPU time of one 512K prefill on Nemotron 3.5 Lightning at $K=L/16$ by cluster size $u$, drawn as Figure~\ref{fig:cluster-size-profile}.}
\label{fig:nemotron-size-profile}
\end{figure}

Figure~\ref{fig:nemotron-size-profile} breaks down prefill time by cluster size.
Smaller clusters mainly increase the time spent on attention over selected blocks, gathering, and merging, but the increase is smaller than on the Qwen models.

\begin{table}[t]
\caption{The values behind the bottom row of Figure~\ref{fig:grid-qwen}, Nemotron 3.5 Lightning at 512K.}
\label{tab:nemotron-grid}
\centering
\small
\setlength{\tabcolsep}{2.6pt}
\begin{tabular}{r rrrr rrrr rrrr}
\toprule
& \multicolumn{4}{c}{MK-3 (\%) at $K$} & \multicolumn{4}{c}{MK-2 (\%) at $K$} & \multicolumn{4}{c}{Prefill (s) at $K$} \\
\cmidrule(lr){2-5} \cmidrule(lr){6-9} \cmidrule(lr){10-13}
$u$ & 8K & 16K & 32K & 64K & 8K & 16K & 32K & 64K & 8K & 16K & 32K & 64K \\
\midrule
Dense & \multicolumn{4}{c}{100.0} & \multicolumn{4}{c}{99.0} & \multicolumn{4}{c}{35.72} \\
4{,}096 & 3.0 & 19.0 & 60.0 & 89.0 & 6.0 & 14.0 & 32.0 & 62.0 & 11.30 & 12.46 & 14.80 & 19.96 \\
2{,}048 & 11.0 & 35.0 & 84.0 & 90.0 & 7.0 & 20.0 & 49.0 & 81.0 & 11.98 & 13.26 & 15.72 & 20.75 \\
1{,}024 & 25.0 & 65.0 & 87.0 & 94.0 & 17.0 & 33.0 & 72.0 & 91.0 & 12.95 & 14.55 & 17.21 & 22.24 \\
512 & 56.0 & 79.0 & 93.0 & 97.0 & 34.0 & 62.0 & 83.0 & 95.0 & 14.81 & 16.83 & 20.09 & 25.09 \\
256 & 81.0 & 86.0 & 93.0 & 99.0 & 62.0 & 78.0 & 92.0 & 98.0 & 19.86 & 22.60 & 26.64 & 31.29 \\
128 & 82.0 & 89.0 & 97.0 & 100.0 & 68.0 & 87.0 & 96.0 & 96.0 & 30.10 & 34.25 & 40.17 & 44.82 \\
\bottomrule
\end{tabular}

\end{table}

Table~\ref{tab:ruler-by-task} reports all five RULER-5 tasks at 512K alongside the Qwen models.
Unlike the Qwen models, Nemotron loses substantial accuracy on multi-value retrieval as well as multi-key retrieval.
Doubling the budget recovers much of this loss while retaining prefill speedups of $1.5$--$2.0\times$ from 256K to 1M (Table~\ref{tab:nemotron}).
The budget increase has a smaller effect on decode time.

\clearpage
\section{Prior sparse-attention methods}
\label{app:related-table}
\suppressfloats[t]

Table~\ref{tab:related-methods} summarizes selected sparse-attention methods and their evaluations.
Here, hybrid models combine full attention with linear-attention or state-space layers.
SpotAttention keeps prefill dense and evaluates accuracy on hybrid Qwen3.5 models~\citep{ahmad2026spotattention}.
It reports decode throughput on Qwen3-8B.
UniPrefill also evaluates several training-free prefill methods on hybrid models up to 128K tokens~\citep{fan2026uniprefill}.

\begin{table}[h]
\centering
\small
\setlength{\tabcolsep}{4pt}
\caption{Sparse-attention methods as their papers describe and evaluate them.
The groups are training-free methods aimed mainly at prefill, training-free methods aimed mainly at decoding, and methods with learned components.
Learned components are those trained for sparsity beyond the pretrained model.
Whole-model time means that the paper times the whole model, such as end-to-end latency, time to first token, or time per output token, rather than only the attention layer or kernel.
DSA reports token costs derived from deployed-service benchmarks.}
\label{tab:related-methods}
\begin{tabularx}{\linewidth}{>{\raggedright\arraybackslash}X >{\raggedright\arraybackslash}p{2.8cm} >{\raggedright\arraybackslash}p{2.4cm} >{\raggedright\arraybackslash}p{1.2cm} >{\raggedright\arraybackslash}p{1.6cm}}
\toprule
Method & Learned components & Sparse phases & Hybrid model & Whole-model time \\
\midrule
MInference~\citep{jiang2024minference} & None & Prefill & No & Yes \\
FlexPrefill~\citep{lai2025flexprefill} & None & Prefill & No & No \\
FlashPrefill / V2~\citep{fan2026flashprefillv2,fan2026flashprefill} & None & Prefill & No & Yes \\
QuoKA~\citep{jones2026quoka} & None & Prefill, decode & No & Yes \\
\addlinespace[4pt]
Quest~\citep{tang2024quest} & None & Decode & No & Yes \\
Squeezed~\citep{hooper2025squeezed} & None & Prefill, decode & No & No \\
ClusterKV~\citep{liu2025clusterkv} & None & Decode & No & Yes \\
Tactic~\citep{zhu2026tactic} & None & Decode & No & Yes \\
\addlinespace[4pt]
SAAP~\citep{mazare2025saap} & Query classifier & Decode & No & No \\
SpotAttention~\citep{ahmad2026spotattention} & Indexer & Decode & Yes & Yes \\
RTPurbo~\citep{zhou2026rtpurbo} & Indexer and model & Prefill, decode & No & No \\
MoBA / FlashMoBA~\citep{lu2025moba,xiao2025flashmoba} & Model & Prefill & No & No \\
DSA~\citep{deepseek2025v32} & Indexer and model & Prefill, decode & No & Service cost \\
MSA~\citep{lai2026msa} & Indexer and model & Prefill, decode & No & No \\
QSA~\citep{qwen2026qsa} & Indexer and model & Prefill, decode & Yes & No \\
HISA~\citep{xu2026hisa} & Indexer and model & Prefill, decode & No & Yes \\
\midrule
This work & None & Prefill, decode & Yes & Yes \\
\bottomrule
\end{tabularx}
\end{table}

ClusterKV and Tactic build a key index during dense prefill for subsequent decoding~\citep{liu2025clusterkv,zhu2026tactic}.
Squeezed builds its index offline for a fixed context~\citep{hooper2025squeezed}.
Our method builds an index within each request and uses it for the prompt's own prefill.
Our corpus-fitted centroids in Section~\ref{sec:results-selection} are scored with the model's queries, whereas SAAP learns a query classifier~\citep{mazare2025saap}.

BLASST skips softmax and attention--value computation for low-scoring blocks~\citep{yuan2025blasst}.
Sketch\&Walk combines sketched scores across layers to select blocks~\citep{le2026sketchwalk}.
Kascade and IndexCache reuse token selections across layers~\citep{deshmukh2025kascade,bai2026indexcache}.

\clearpage
\section{Examples of cluster members}
\label{app:cluster-examples}
\suppressfloats[t]

Figure~\ref{fig:cluster-examples} shows one cluster from an MRCR prompt and one from an MK-3 prompt.
Each highlighted span is a single member token, with surrounding text for context.
Cosine similarity is measured between the normalized pre-RoPE key and its assigned centroid.
Each cluster comes from a partition of 256 clusters fitted to the unmodified model's keys.
Layer, KV-head, and token-position indices are zero-based.
The figure shows the three nearest and three farthest members of each cluster among the saved examples.

The MRCR cluster groups environmental vocabulary across several passages.
Its lower-similarity members extend to medicine, history, and mathematics.
For MK-3, the nearest members occur at the beginning of key identifiers, but lower-similarity members also include the beginning of a value identifier.

In two MK-3 and two MRCR prompts, about 70\% of MK-3 clusters and 90\% of MRCR clusters have no single token type accounting for more than half their members.
These counts use all cluster members across eight layers and four KV heads per prompt.

\subsection{Further clusters}
\label{app:additional-clusters}

\begin{figure}[t]
\centering
\includegraphics[width=\linewidth]{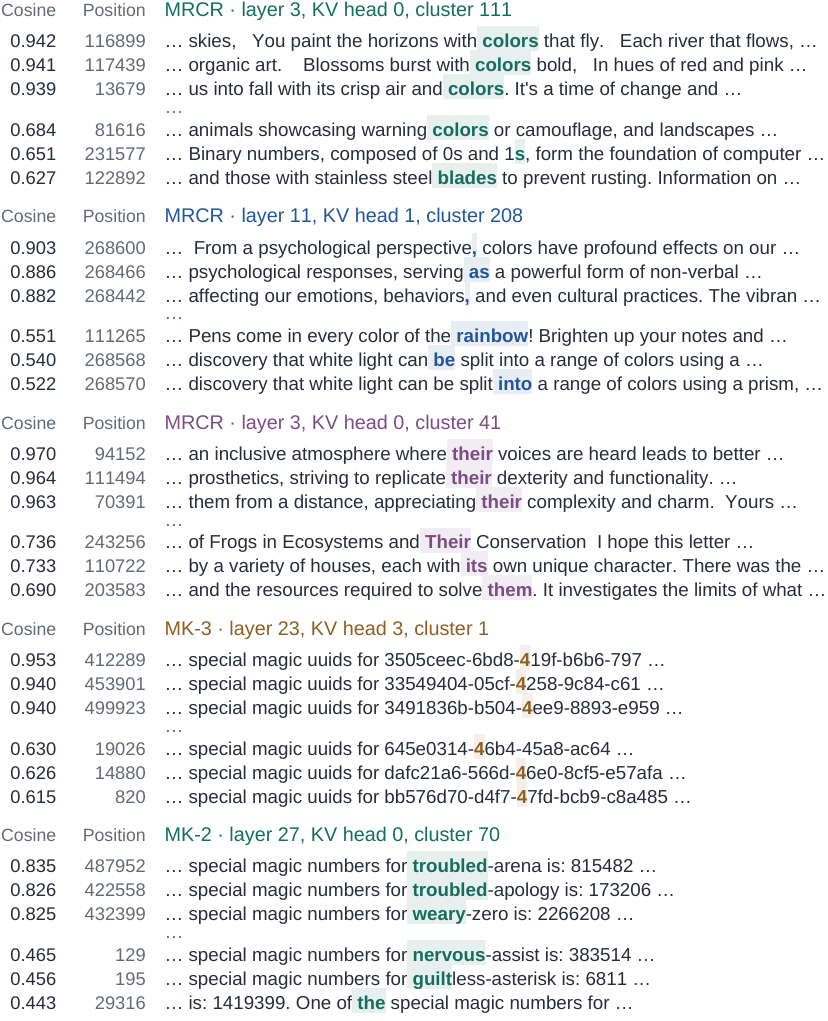}
\caption{Five additional clusters containing color vocabulary, tokens in a passage about color, pronouns, UUID digits, and adjectives in MK-2 keys.
Positions are token indices, and highlights mark member tokens.
Rows are sorted by key--centroid cosine similarity, with a gap between the two ends.}
\label{fig:additional-clusters}
\end{figure}

Figure~\ref{fig:additional-clusters} shows five further clusters, using three members from each end of the saved cosine range.
The first and third clusters use the same MRCR prompt as Figure~\ref{fig:cluster-examples}, the second and fourth use separate MRCR and MK-3 prompts, and the fifth uses an MK-2 prompt.

The first cluster groups occurrences of \emph{colors} across descriptions, while its lower-similarity members extend to other words.
The second includes punctuation and function words within a passage about color perception.
The third groups pronouns and possessives across different topics.
In the fourth, the nearest members are the digit 4 at the start of the third group of a key UUID.
One lower-similarity member instead occurs in the second group.
The fifth comes from an MK-2 prompt with lines such as ``One of the special magic numbers for troubled-arena is: 8154823.''
All ten saved nearest members of this cluster are negative adjectives in different keys, such as \emph{troubled}, \emph{weary}, \emph{abusive}, \emph{disgusted}, and \emph{ashamed}.
Its lowest-similarity members include other tokens of the same lines.

\end{document}